\documentclass[3p,times]{elsarticle}

\usepackage{ecrc}

\volume{00}

\firstpage{1}

\journalname{Procedia Computer Science}

\runauth{}

\jid{procs}

\jnltitlelogo{Procedia Computer Science}

\CopyrightLine{2011}{Published by Elsevier Ltd.}

\usepackage[linesnumbered,ruled]{algorithm2e}
\usepackage{amsmath,amssymb,amsfonts}
\usepackage{graphicx}
\usepackage{textcomp}
\usepackage{xcolor}
\usepackage{bm}
\usepackage{multirow}
\usepackage{booktabs}
\UseRawInputEncoding

\usepackage[figuresright]{rotating}

\begin{document}

\begin{frontmatter}



\dochead{}

\title{Cooperative coevolutionary Modified Differential Evolution with Distance-based Selection for Large-Scale Optimization Problems in noisy environments through an automatic Random Grouping}

\author[ad:1]{Rui Zhong\corref{1}\fnref{1}}
\address[ad:1]{Graduate School of Information Science and Technology, Hokkaido University}

\author[ad:2]{Masaharu Munetomo\corref{1}}
\address[ad:2]{Information Initiative Center, Hokkaido University}

\begin{abstract}
Many optimization problems suffer from noise, and nonlinearity check-based decomposition methods (e.g. Differential Grouping) will completely fail to detect the interactions between variables in multiplicative noisy environments. Thus, it is difficult to decompose large-scale optimization problems (LSOPs) in noisy environments. In this paper, we propose an automatic Random Grouping (aRG), which does not need any explicit hyperparameter specified by users. Simulation experiments and mathematical analysis show that aRG can detect the interactions between variables without the fitness landscape knowledge, although the probabilistic analysis of aRG is inferior to the conventional Random Grouping\cite{Yang:08}, the sub-problems decomposed by aRG have smaller scales, which can alleviate the negative influence of the curse of dimensionality and accelerate the optimization of EAs. Based on the cooperative coevolution (CC) framework, we introduce an advanced optimizer named Modified Differential Evolution with Distance-based Selection (MDE-DS) to enhance the search ability in noisy environments. Compared with canonical DE, the parameter self-adaptation, the balance between diversification and intensification, and the distance-based probability selection endow MDE-DS with stronger ability in exploration and exploitation. To evaluate the performance of our proposal, we design $500$-D and $1000$-D problems with various separability in noisy environments based on the CEC2013 LSGO Suite. Numerical experiments show that our proposal has broad prospects to solve LSOPs in noisy environments and can be easily extended to higher-dimensional problems.
\end{abstract}

\begin{keyword}
Large-scale optimization problems (LSOPs), Cooperative Coevolution (CC), automatic Random Grouping (aRG), noisy environments, MDE-DS.

\end{keyword}
\end{frontmatter}

\section{Introduction} \label{sec:1}
Evolutionary algorithms (EAs) have been applied to deal with many optimization problems with great success\cite{Thomas:93, Darrell:01}. However, as the dimension of the problem increases, the search space increases exponentially, which conducts the performance of traditional EAs to degrade dramatically. This phenomenon is also known as the curse of dimensionality\cite{Mario:00}. Many research efforts in computer science ranging from computational linear algebra\cite{Martinsson:20} and machine learning\cite{Bottou:18} to numerical optimization\cite{Gould:05} have been published to alleviate the curse of dimensionality. 

In a large number of optimization problems, the presence of noise when sampling is also a knotty problem. Noise will make a great difference between the observation fitness landscape and the real fitness landscape, and the existence of noise will mislead the direction of optimization. Thus, it is difficult to solve the optimization problem in noisy environments with traditional EAs. Many researchers focus on the low-dimensional optimization problems in noisy environments and proposed many efficient methods such as explicit averaging\cite{Laura:95}, implicit averaging\cite{Diaz:15}, fitness estimation\cite{Youhei:15}, and so on. However, most of the existing studies concentrate on relatively low-dimensional problems (up to $100$-D), and problems in noisy environments with hundreds of decision variables have rarely been studied. In practice, many noisy optimization problems are high-dimensional, such as parameters and structures optimization of deep neural networks\cite{Chen:20} and subset selection\cite{Qian:17}. 

The combination of large-scale features and noise makes the difficulty of solving the problem explosive. Cooperative Coevolution (CC)\cite{Potter:94} is a simple and efficient framework to solve LSOPs. CC decomposes the original high-dimensional optimization problem into many low-dimensional problems and solves them alternately, which makes great success in solving large-scale continuous\cite{Sun:17}, multi-objective\cite{Ma:16}, constraint optimization problems\cite{Sayed:15}, real-world problems\cite{Mei:14}, and so on.

This paper is based on the CC framework to solve large-scale optimization problems (LSOPs) in noisy environments. The key to the successful implementation of CC is the design of the variable grouping strategy. In theory, a perfect grouping strategy can exponentially reduce the search space without losing optimization accuracy and alleviate the negative effects of the curse of dimensionality, while a poor grouping strategy will mislead the direction of optimization and guide the optimization into a dilemma. Therefore, identifying the separability between variables and designing high-accuracy variable grouping methods have become a hot research topic in the past years. Taking the Linkage Identification by Nonlinearity Check for Real-Coded GA (LINC-R)\cite{Tezuka:04} as the pioneer, grouping methods such as DG\cite{Omidvar:14}, DG2\cite{Omidvar:17}, XDG\cite{Sun:15}, and graphDG\cite{Ling:16} are derived for high-dimensional problems, which are considered high-accuracy grouping methods. These methods detect the interactions based on fitness differences calculated by perturbations. Although these methods are high accuracy in interaction identification, the implementations are environmentally demanding, which requires the error between the observed objective value and the real objective value should be infinitely close to 0, the specific challenge will be explained in section \ref{sec:2.3}. Therefore, these clustering grouping methods face unprecedented difficulty in noisy environments, although it is possible to identify the interactions through these grouping methods with the parameter modification in additive noise, it is not an easy task\cite{Wu:22}.

In this paper, we propose an automatic Random Grouping (aRG) for LSOPs with multiplicative noise. Unlike many previous studies on Random Grouping, our proposal does not need any explicit hyperparameter provided by users, and all grouping process is completely implemented by the grouping strategy itself. And we introduce the advanced optimizer for optimization problems in noisy environments named Modified Differential Evolution with Distance-based Selection (MDE-DS)\cite{Ghosh:17} as the basic optimizer (MDE-DSCC-aRG). The main contributions of this paper are as follows.

(1). This paper proposes an automatic Random Grouping for optimization in noisy environments. Simulation experiments and mathematical analysis show that this simple strategy can also divide the interacting variables into the same group in multiple trial runs when there is no information or only noisy information about the fitness landscape can be obtained.

(2). MDE-DS is applied as the optimizer for sub-problems and is well-performed on various benchmark functions in noisy environments. The results in this paper further demonstrate that MDE-DS can accelerate cooperative coevolutionary optimization significantly. 

(3). Numerical experiments demonstrate that our proposal is competitive with some state-of-the-art grouping methods for LSOPs in noisy environments. To the best of our knowledge, not much work has been reported on adopting grouping methods for CC in noisy environments.

The remainder of this paper is organized as follows. Section \ref{sec:2} covers preliminaries, the modified DE with distance-based selection, the challenge of nonlinearity check working in noisy environments, and studies for anti-noise in optimization problems. Section \ref{sec:3} introduce our proposal, MDE-DSCC-aRG. Experiments and analyses of MDE-DSCC-aRGfs behavior and performance are described in section \ref{sec:4}. In section \ref{sec:5}, we discuss some open topic for future research. Finally, Section \ref{sec:6} concludes this paper.

\section{Preliminaries and related works} \label{sec:2}
Many variable grouping methods based on CC have been published in recent years, but only a few proposals focus on LSOPs in noisy environments. In this section, we introduce some preliminaries including the variable interactions, the noisy environments, and the principle of variable grouping methods. Second, an advanced optimizer MDE-DS is described in detail, then, the challenge of traditional variable grouping methods working in noisy environments is covered. Finally, anti-noise strategies for optimization problems are briefly reviewed. 

\subsection{Preliminaries} \label{sec:2.1}

\subsubsection{Variable interactions} \label{sec:2.1.1}
The concept of variable interactions is generated from biology. If a phenotypic feature is related to two or more genes, then these genes are considered with interactions. Goldberg\cite{Goldberg:96} extended this to optimization problems. When identifying the interaction between $x$ and $y$, if $\frac{\partial^2 f}{\partial x\partial y}=0$, then there is no interaction between $x$ and $y$. Notice that the presence of variable interaction between $x$ and $y$ is not completely equivalent to nonseparability. Take a simple example, $f(x)=(x_1+x_2)^2$, $x_i \in [0,10]$, although $x_1$ and $x_2$ interact, $x_1$ and $x_2$ are still separable in limited search space considering monotonicity. Thus, the ultimate goal of grouping methods is to identify the separability between variables and divide the variables into groups properly. 

\subsubsection{Noisy environments} \label{sec:2.1.2}
Additive noise and multiplicative noise are two kinds of representative noises, and they also widely exist in practical problems. Mathematically, the noisy objective function $f^N(x)$ of a trial solution ${\rm X}$ can be defined:
\begin{equation}
	\begin{aligned}
		\label{eq:1}
		f^{N}({\rm X}) = f({\rm X}) + \eta 
	\end{aligned}
\end{equation}
\begin{equation}
	\begin{aligned}
		\label{eq:2}
		f^{N}({\rm X}) = f({\rm X}) \cdot (1+\beta)
	\end{aligned}
\end{equation}
where $f(x)$ is the real objective function, Eq (\ref{eq:1}) is $f(x)$ in addictive noisy environments, and Eq (\ref{eq:2}) reveals the relationship between $f(x)$ and $f^N (x)$ in multiplicative noisy environments. $\eta$ and $\beta$ are random noises (such as Gaussian noise). In this paper, we focus on optimization problems in multiplicative noisy environments.

\subsubsection{The principle of variable grouping methods} \label{sec:2.1.3}
CC framework was proposed in 1994\cite{Potter:94} and become a mature and efficient framework for solving LSOPs. A variable grouping strategy is necessary for the employment of CC, and sub-problems are optimized alternately. Since the candidate solutions of each sub-problem cannot form a complete solution, representative solutions of other sub-problems are necessary to form a complete solution for evaluation. The context vector is composed of these representative solutions and updated iteratively and acts as the context in which the cooperation occurs. Generally, there are mainly three principles of variable grouping methods: automatic, semi-automatic, and $m \times k$-strategy.

(1). Automatic: In automatic variable grouping methods, the formation of sub-problems completely depends on the algorithm process and built-in parameters. Representative methods include nonlinearity check and monotonicity detection and derive LINC-R\cite{Tezuka:04}, DG\cite{Omidvar:14}, DG2\cite{Omidvar:17} and LIMD\cite{Munetomo:99}, CCVIL\cite{Chen:10} respectively. These class methods are considered high-accuracy and expensive grouping methods in noiseless environments 

(2). Semi-automatic: In semi-automatic variable grouping methods, the size or the number of sub-problems is required to be specified by users. The typical methods such as AVP2\cite{Singh:10} and 4CDE\cite{Rojas:11} employ the statistical methods to form the groups by a threshold or a set of intervals defined on the correlation coefficient. A special semi-automatic grouping is called multilevel\cite{Ke:08} in which the user specifies a list of potential sub-problem sizes for the algorithm to choose from, if there is no improvement in a generation of optimization, the algorithm will randomly choose the candidate sub-problem size based on probability and reconstruct the groups.

(3). $m \times k$-strategy: These methods only need little or even no information about the fitness landscape and require the hyperparameters of group size and the number of groups provided by users to form the sub-problems. The typical methods include Random Grouping\cite{Yang:08, Song:17}, Delta Grouping\cite{Omidvar:10}, Fitness Difference Partitioning\cite{Dai:16}, and so on. Although little information is known about the fitness landscape of these methods, a rigorous mathematical proof states that, even in Random Grouping, the probabilities of two interacting variables being divided into the same group at least once or twice in multiple cycles are quite high.

\subsection{Modified DE with Distance-based Selection} \label{sec:2.2}
Differential Evolution Algorithm (DE)\cite{Storn:96} was first proposed in 1995 and has been wildly applied in data mining\cite{He:09}, pattern recognition\cite{Du:07}, artificial neural networks\cite{Slowik:08}, and other fields due to its characteristics such as easy implementation, fast convergence speed, and strong robustness. MDE-DS is\cite{Ghosh:17} designed for continuous optimization problems in presence of noise with the modification in mutation, crossover, and selection, and the detailed description of MDE-DS is as follows. 

\subsubsection{Parameter control} \label{sec:2.2.1}
The constant $F$ and $Cr$ are unnecessary, as $F$ is randomly sampled from $0.5$ to $2$ for each mutation operation and $Cr$ is randomly switched between 0.3 to 1 for each target vector. Switching $F$ between two extreme corners of the feasible range is conducive to attaining a balance between exploration and exploitation of the search. And there is a new parameter $b$ (the blending rate) in blending crossover, whose value is also randomly chosen from among three candidates: a low value of $0.1$, a medium value of $0.5$, and a high value of $0.9$. The utility of such a switching scheme has been discussed in paper \cite{Arka:17} for solving LSOPs. 

\subsubsection{Mutation} \label{sec:2.2.2}
MDE-DS includes two different mutation strategy and switches them randomly with $50\%$ probability. 

In the population centrality-based mutation, the elite subpopulation (top 50\%) is selected and $\widetilde{\vec{X}}_{best, G}$ is calculated by the arithmetic mean (centroid) of the subpopulation individuals. Eq (\ref{eq:3}) is adopted to mutate the $i^{th}$ individual. 
\begin{equation}
	\begin{aligned}
		\label{eq:3}
		\vec{V}_{i, G}=\vec{X}_{r1, G}+F\left(\widetilde{\vec{X}}_{best, G} - \vec{X}_{r2, G}\right)
	\end{aligned}
\end{equation}
where $\vec{X}_{r1, G}$ and $\vec{X}_{r2, G}$ are two different individuals corresponding to randomly chosen indices $r1$ and $r2$. $\vec{V}_{i, G}$ is the newly generated mutant vector corresponding to the current target vector for present generation $G$. 

In the DMP-based mutation scheme, the best individual $\vec{X}_{best, G}$ in each generation is selected and the dimension-wise average is implemented for both $\vec{X}_{best, G}$ and the current target individual $\vec{X}_{i, G}$. The mutation is generated in the following way: 
\begin{equation}
	\begin{aligned}
		\label{eq:4}
		\vec{V}_{i, G}=\vec{X}_{i, G}+\Delta_{m} \cdot \left(  \frac{\vec{M}_{i, G}}{\Vert \vec{M}_{i, G} \Vert}  \right)
	\end{aligned}
\end{equation}
where $\Delta_{m}=(X_{best_{dim}, G}-X_{i_{dim}, G})$, with $X_{best_{dim}, G}=\frac{1}{D}\sum_{k=1}^{D}x_{best_k, G}$ and $X_{i_{dim}, G}=\frac{1}{D}\sum_{k=1}^{D}x_{i_k, G}$. $\frac{\vec{M}_{i, G}}{\Vert \vec{M}_{i, G} \Vert}$ is a unit vector with random direction. 

The significance of the population centrality-based mutation scheme is that it balances greediness while still maintaining some level of diversity. For example, it is less greedy than the DE/best/1 scheme and hence the probability of the optimization trapped in local optima is less. On the other hand, the DMP-based mutation scheme prefers exploration\cite{Kundu:13}, and thus, in absence of any feedback about the nature of the function, an unbiased combination of these two methods is applied.

\subsubsection{Crossover} \label{sec:2.2.3}
Crossover plays an important role in generating promising offspring from two or more existing individuals within the function landscape. blending crossover is employed in MDE-DS and described in Eq (\ref{eq:5})
\begin{equation}
	\label{eq:5}
	\begin{aligned}
	u_{j,i,G}=
	\begin{cases}
		b \cdot x_{j,i,G} + (1-b) \cdot v_{j,i,G}  \\
		x_{j,i,G} 
	\end{cases}
	\end{aligned}
\end{equation}
where $u_{j,i,G}$ and $v_{j,i,G}$ are the $j^{th}$ dimensions of the trial and donor vectors respectively corresponding to current index $i$ in generation $G$ and $x_{j,i,G}$ is the $j^{th}$ dimension of the current population individual $\vec{X}_{i, G}$. Blending recombination has one parameter $b$, which is randomly selected from 0.1, 0.5, and 0.9. The concrete analysis can be referred in \cite{Ghosh:17}. 

\subsubsection{Selection} \label{sec:2.2.4}
The canonical DE selects the offspring based on a simple greedy strategy. However, if the fitness landscape gets corrupted with noise, the greedy selection suffers a lot because in this case the original fitness of parent and offspring is unknown and it can be well nigh impossible to infer when an offspring is superior or inferior to its parent. Thus, the design of selection is the key to anti-noise. To handle the presence of noise, a novel distance-based selection mechanism is introduced without any extra parameter. There are three cases of the proposed selection mechanism which are described subsequently:
\begin{equation}
	\label{eq:6}
	\begin{aligned}
		\vec{X}_{i,G+1}=
		\begin{cases}
			\vec{U}_{i,G}, \ if \ \frac{f(\vec{U}_{i,G})}{f(\vec{X}_{i,G})} \leq 1  \\
			\vec{U}_{i,G}, \ if \ \frac{f(\vec{U}_{i,G})}{f(\vec{X}_{i,G})} > 1 \ and \ p_s \leq e^{-\frac{\Delta f}{Dis}} \\
			\vec{X}_{i,G}, \ else
		\end{cases}
	\end{aligned}
\end{equation}
In case 1, when $\frac{f(\vec{U}_{i,G})}{f(\vec{X}_{i,G})} \leq 1$, the offspring replaces the parent and survives to the next generation.

In case 2, although the parent performs better than the offspring, the offspring still can be preserved and survive into the next generation based on a stochastic principle. And the probability is calculated by $e^{-\frac{\Delta f}{Dis}}$, where $\Delta f=\left \arrowvert f(\vec{U}_{i,G}) - f(\vec{X}_{i,G}) \right\arrowvert$ represents the absolute fitness difference between $\vec{U}_{i,G}$ and $\vec{X}_{i,G}$, $Dis=\sum_{k=1}^{D}\left \arrowvert u_{i,k}-x_{i,k} \right\arrowvert$ is the Manhattan distance between those two vectors. Manhattan distance is applied because of its simplicity and computational efficiency, and $p_s$ is a random number generated from 0 to 1.

In case 3, If the parent significantly outperforms than offspring, then the offspring is removed and the parent persists to the next generation.

\begin{figure}[htb]
	\centering
	\includegraphics[width=12cm]{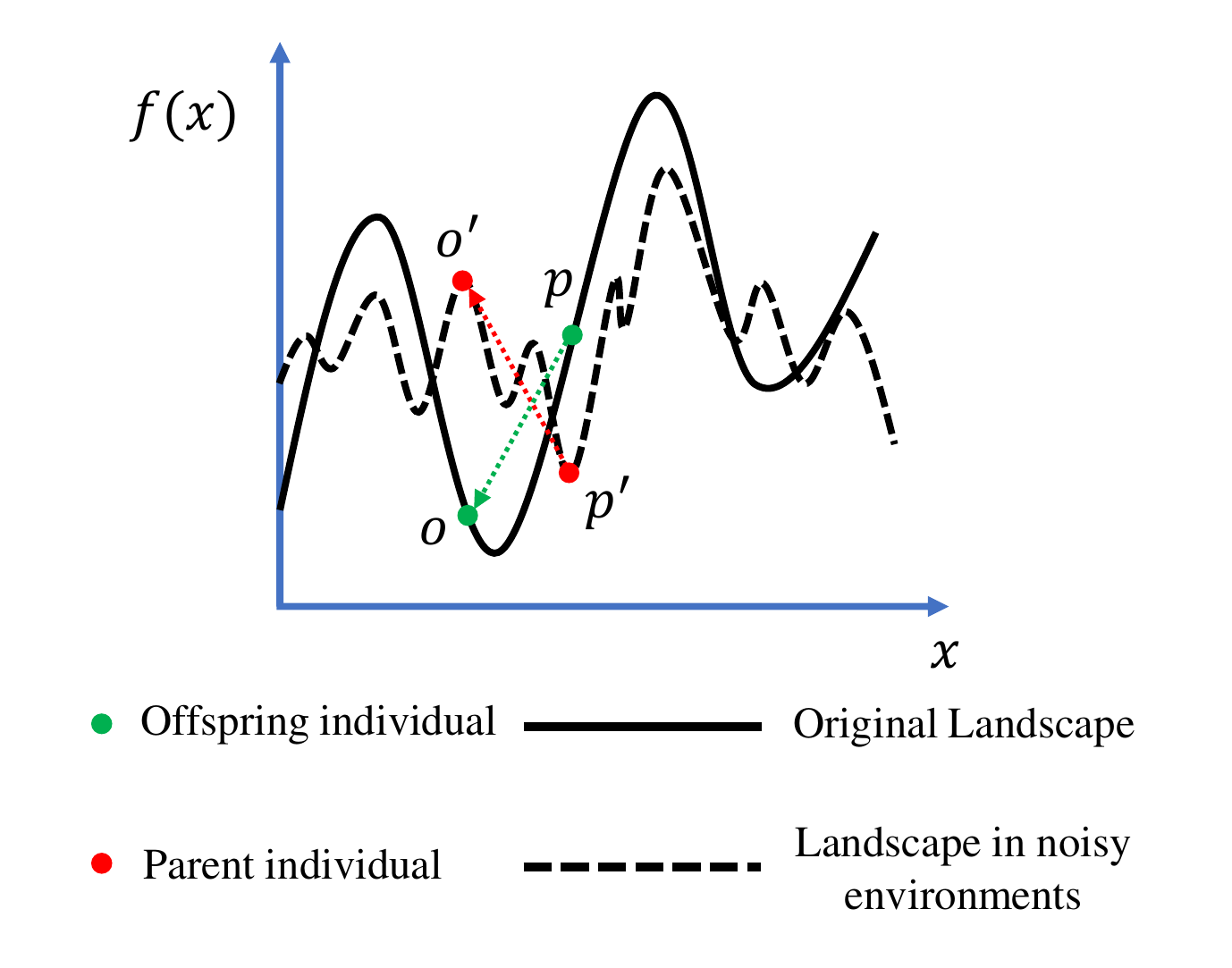}
	\caption{A selection works on fitness landscape in noisy environments}
	\label{fig:1}
\end{figure}

This selection process is further illustrated in Fig. \ref{fig:1}.
Fig. \ref{fig:1} shows a fitness landscape scenario both in noiseless environments and noisy environments, $p$ and $p^{'}$ represent the parent individual in the original fitness landscape and landscape in noisy environments, $o$ and $o^{'}$ represent the offspring individual in original fitness landscape and landscape in noisy environments respectively. The fitness information we can observe is only in noisy environments, so in minimization problems, $o^{'}$ will be rejected to replace the $p^{'}$ in the next generation. The objective value of $p$ is better than $o$ in the real fitness landscape, and if we re-evaluate the $o^{'}$ and $p^{'}$, the domination may be changed. The mechanism of selection in MDE-DS allows the algorithm to give us some probabilistic flexibility to select worse solutions as in noise-affected landscapes.

In summary, the Pseudocode of MDE-DS is shown in Algorithm \ref{alg:1}
\begin{algorithm}
	\label{alg:1}
	\DontPrintSemicolon
	\SetAlgoLined
	\KwIn {${\rm Population \ size}:s;{\rm Dimension}:D;{\rm Max generation}:G; {\rm Search \ space}: lb, ub$}
	\KwOut {${\rm Optimum}: x_{best}$}
	\SetKwFunction{FDE}{\textbf{MDE-DS}}
	\SetKwProg{Fn}{Function}{:}{}
	\Fn{\FDE{$s, D, G, lb, ub$}}{
		$t \gets 0$\;
		$\blacktriangleright$ (Initialization) \;
		\For{$i=0 \ to \ s$}{
			\For{$j=0 \ to \ D$}{
				$X_{j,i,t} \gets lb_j+ \textbf{rand}(0, 1) \times (ub_j - lb_j)$\;
			}
		}
		$FP \gets \textbf{evaluate}(X)$ \;
		$x_{best} \gets \textbf{bestIndividual}(FP, X)$ \;
		\While{$t < G$ \ {\rm and not stop criterion}}{
			\For{$i=0 \ to \ s$}{
				$\blacktriangleright$ (Mutation) \;
				$r \gets \textbf{rand}(0, 1)$ \;
				\If{$r \leq 0.5$}{
					$EP_G \gets \textbf{elitePopulation}(X, F, 0.5)$ // extract top 50\% subpopulation\; 
					$\widetilde{\vec{X}}_{best, G} \gets \frac{2}{s}\sum_{k=1}^{s/2}EP_{k, G}$ \;
					$F \gets \textbf{rand}(0.5, 2)$ \;
					$\vec{V}_{i, G}=\vec{X}_{r1, G}+F\left(\widetilde{\vec{X}}_{best, G} - \vec{X}_{r2, G}\right)$ \;
				} \Else{
					$X_{best_{dim}, G} \gets \frac{1}{D}\sum_{k=1}^{D}x_{best_k}$ \;
					$X_{i_{dim}, G} \gets \frac{1}{D}\sum_{k=1}^{D}x_{i_k}$ \;
					$\vec{M}_{i, G} \gets \textbf{unitVector}()$ \;
					$\vec{V}_{i, G}=\vec{X}_{i, G}+\Delta_{m} \cdot \left(  \frac{\vec{M}_{i, G}}{\Vert \vec{M}_{i, G} \Vert}  \right)$ \;
				}
				$\blacktriangleright$ (Crossover) \;
				$Cr \gets \textbf{rand}(0.3, 1)$ \;
				$b \gets \textbf{randChoice}([0.1, 0.5, 0.9])$ \;
				\For{$j=0 \ to \ D$}{
					$u_{j,i,G}=
					\begin{cases}
						b \cdot x_{j,i,G} + (1-b) \cdot v_{j,i,G}  \\
						x_{j,i,G} 
					\end{cases}$ \;
				}
				$\blacktriangleright$ (Selection) \;
				$FP_i \gets \textbf{evaluate}(U_i)$ \;
				$\Delta f_i=\left \arrowvert f(\vec{U}_{i}) - f(\vec{X}_{i}) \right\arrowvert$ \;
				$Dis=\sum_{k=1}^{D}\left \arrowvert u_{i,k}-x_{i,k} \right\arrowvert$ \;
				$p_s \gets \textbf{rand}(0, 1)$ \;
				$\vec{X}_{i,G+1}=
				\begin{cases}
					\vec{U}_{i,G}, \ if \ \frac{f(\vec{U}_{i,G})}{f(\vec{X}_{i,G})} \leq 1  \\
					\vec{U}_{i,G}, \ if \ \frac{f(\vec{U}_{i,G})}{f(\vec{X}_{i,G})} > 1 \ and \ p_s \leq e^{-\frac{\Delta f}{Dis}} \\
					\vec{X}_{i,G}, \ else
				\end{cases}$ \;
			}
				$t \gets t+1 $\;
		}
		$\textbf{return} \ x_{best}$
	}
	\caption{MDE-DS}
\end{algorithm}

\subsection{The challenge of nonlinearity check in noisy environments}  \label{sec:2.3}
As we describe in section \ref{sec:1}, nonlinearity check-based variable grouping methods, although considered high accuracy in noiseless environments, face an unprecedented challenge for optimization problems in noisy environments. Here, we take the LINC-R as an example:
\begin{equation}
	\begin{aligned}
		\label{eq:7}
		\forall s\in Pop:\\
		\Delta_{1}=f^{N}(s_{i})-f^{N}(s)=f(s_{i})(1+\beta_{1})-f(s)(1+\beta_{2}) \\
		\Delta_{2}=f^{N}(s_{ij})-f^{N}(s_{j})=f(s_{ij})(1+\beta_{3})-f(s_{j})(1+\beta_{4}) \\
		if \ \lvert \Delta_{2} - \Delta_{1} \rvert < \varepsilon \\
		then \ x_{i} \ and \ x_{j} \ are \ separable
	\end{aligned}
\end{equation}
${\beta_{i}}$ is Gaussian noise. In noiseless environments ($\beta_{i}=0$), LINC-R can identify the interactions between variables. However, in noisy environments with multiplicative noise, the probability of Eq (\ref{eq:7}) being satisfied is almost 0, theoretically. In practice, decomposition methods developed on the nonlinearity check will completely fail to determine the interactions in multiplicative noisy environments.

\subsection{Anti-noise strategies in EAs}   \label{sec:2.4}
Many optimization problems suffer from noise, and to perform the optimization under the existence of noise, various anti-noise strategies have been proposed in the literature. Following the classification reported in \cite{Jin:05}, two categories of noise handling methods for EAs can be mainly classified; each category can be divided into two sub-categories:
\begin{itemize}
	\item[$\bullet$] Methods which require an increase in the computational cost \\
	(1). Explicit Averaging Methods \\
	(2). Implicit Averaging Methods 
	\item[$\bullet$] Methods which perform hypotheses about the noise \\
	(1). Averaging by means of approximated models \\
	(2). Modification of the Selection Schemes 
\end{itemize}

Explicit averaging methods consider that re-sampling and re-evaluation can reduce the impact of noise on the fitness landscape. As a matter of fact, increasing the re-evaluation times is equivalent to reducing the variance of the estimated fitness. Thus, ideally, an infinite sample size would reduce to $0$ uncertainties in the fitness estimations.

Implicit averaging states that a larger population allows the evaluations of neighbor solutions and thus the fitness landscape in a certain portion of decision space can be estimated. Paper\cite{Fitzpatrick:88} has shown that a large population size reduces the influence of noise on the optimization process, and paper\cite{Miller:96} has proved that a GA with an infinite population size would be noise-insensitive.

Both explicit and implicit averaging methods consume more fitness evaluation times (FEs) to correct the objective value, which is improper or even unacceptable for LSOPs under the FEs limitation. In order to obtain efficient noise filtering without excessive computational cost, various solutions have been proposed in the literature, such as the introduction of the approximated model\cite{Sano:00}, probability-based selection schemes\cite{Iacca:12}, self-adaptative parameter adjustment\cite{Mininno:10}, and so on. 

In this paper, as we mentioned in section \ref{sec:2.2.4}, MDE-DS modifies the selection scheme based on probability, which is efficient to solve LSOPs in noisy environments, especially under the FEs limitation. 

\section{MDE-DSCC-aRG} \label{sec:3} 
\begin{figure*}[htb]
	\centering
	\includegraphics[width=12cm]{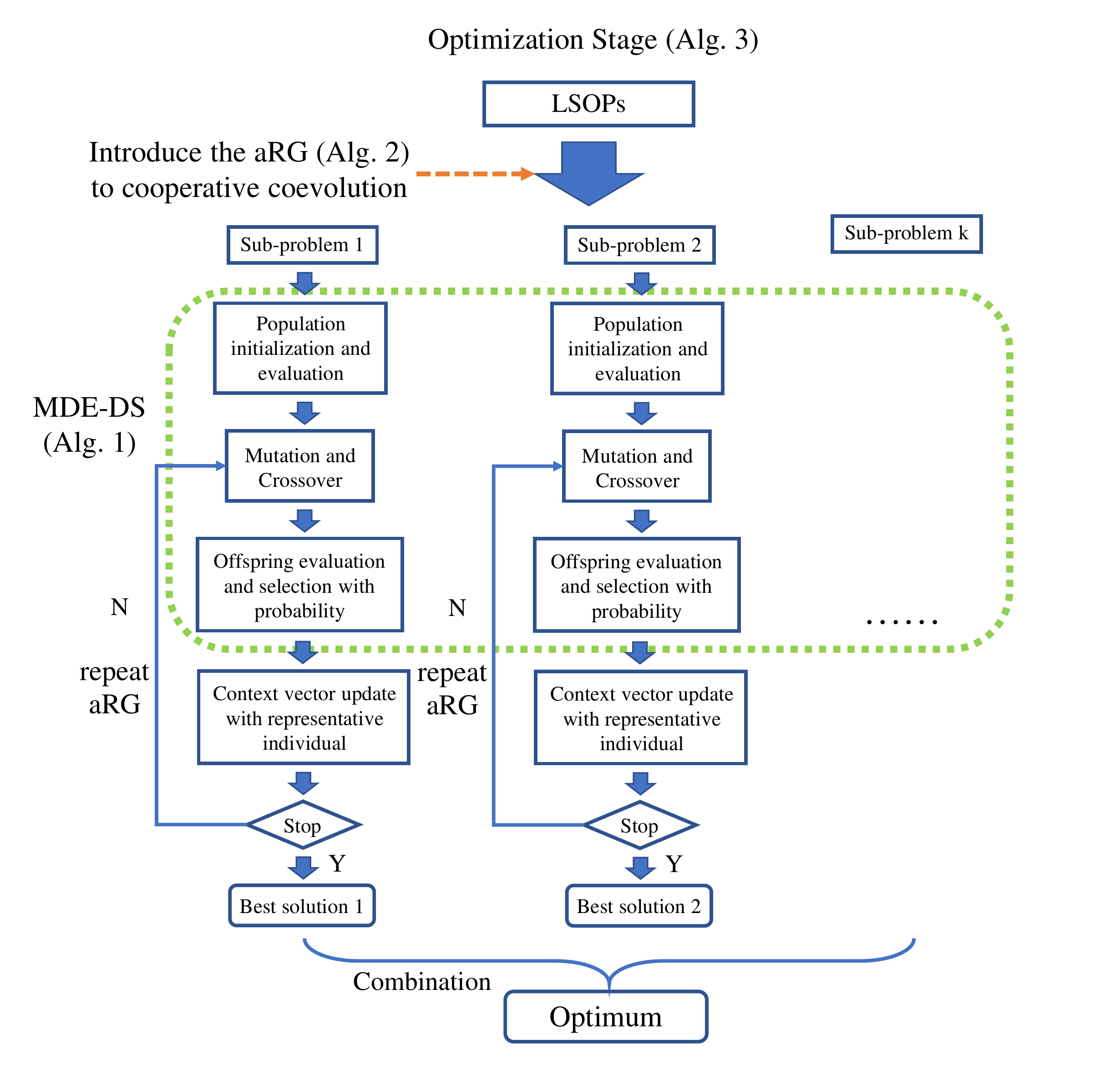}
	\caption{The flowchart of MDE-DSCC-aRG}
	\label{fig:2}
\end{figure*}

In this section, we will introduce the details of our proposal. The flowchart of our proposal is shown in Fig. \ref{fig:2}.

Current popular random grouping methods require the size and the number of sub-problems\cite{Yang:08}, or a candidate list of the sizes of the sub-problems which are specified by users\cite{Ke:08}. Our proposed grouping method is different and more convenient, which is not compulsory for any explicit hyperparameters provided by users. The grouping procedure is completely implemented depending on the logic of the algorithm and the self-adaptive probability. The key steps of our proposed grouping methods can be summarized as follows:

\begin{enumerate}[(1)]
	\setlength{\itemindent}{2em} 
	\item Shuffle the order of the variables and create an empty set to save groups.
	\item Iterate the shuffled variables, and the variable has an equivalent probability to be allocated into existing groups or a new group.
	\item Stop until all variables are accessed, and return the groups.
\end{enumerate}
Algorithm \ref{alg:2} shows the detailed procedure of aRG.
\begin{algorithm}
	\label{alg:2}
	\DontPrintSemicolon
	\SetAlgoLined
	\KwIn {${\rm Dimension}:D$}
	\KwOut {${\rm Groups}: G$}
	\SetKwFunction{FRG}{\textbf{aRG}}
	\SetKwProg{Fn}{Function}{:}{}
	\Fn{\FRG{$D$}}{
		$G \gets \emptyset$\;
		$V \gets [1,2,...,D]$ \;
		$V \gets \textbf{shuffle}(V)$ \;
		\For{$i=0 \ to \ D$}{
			$s \gets \textbf{size}(G)$ \;
			$r \gets \textbf{randint}(0, s)$  \ // Random allocate the variable into a group\; 
			\If{$r==s$}{   
				$G_{r, 0} \gets V_{i}$   \ // Allocated into a new group \;
			} \Else{  
				$k \gets \textbf{size}(G_r)$  \ // Allocated into a exist group \;
				$G_{r, k} \gets V_{i}$ \;
			}
		}
		$\textbf{return} \ G$
	}
	\caption{aRG}
\end{algorithm}
We follow the frequency of the implementation of grouping methods in paper \cite{Nabi:10}, the aRG is repeated at every generation finished. Paper \cite{Yang:08} proved that the probability of two interacting variables $x_i$ and $x_j$ being assigned into the same sub-problem for at least $k$ cycles in DECC-G is
\begin{equation}
	\label{eq:8}
	\begin{aligned}
		P_k=\sum_{r=k}^{N} \binom{N}{r} \left(\frac{1}{m}\right)^r \left(1-\frac{1}{m}\right)^{N-r}
	\end{aligned}
\end{equation}
where $N$ is the total number of cycles and $m$ is the number of sub-problems.  Given $n=1000$, $s=100$, we know that the number of sub-problems would be $m=n/s=10$. If the number of cycles $N=60$, we have: $P_1=0.9982, P_2=0.9862$. Rigorous mathematical proof shows that random grouping is quite effective in capturing variable interactions with little landscape knowledge. 

Our aRG also adopts the strategy of random grouping, but considering the higher randomness of aRG, each grouping result has a different number of groups, and the size of the sub-problems is also different. Therefore, the simulation experiment is implemented and the statistical method is applied to approximately estimate the probability of two interacting variables $x_i$ and $x_j$ divided at least once or twice into $k$ rounds.  

First, we run aRG $100,000$ times and count the number of sub-problems occurring, which is shown in Fig. \ref{eq:3}.
\begin{figure*}[htb]
	\centering
	\includegraphics[width=12cm]{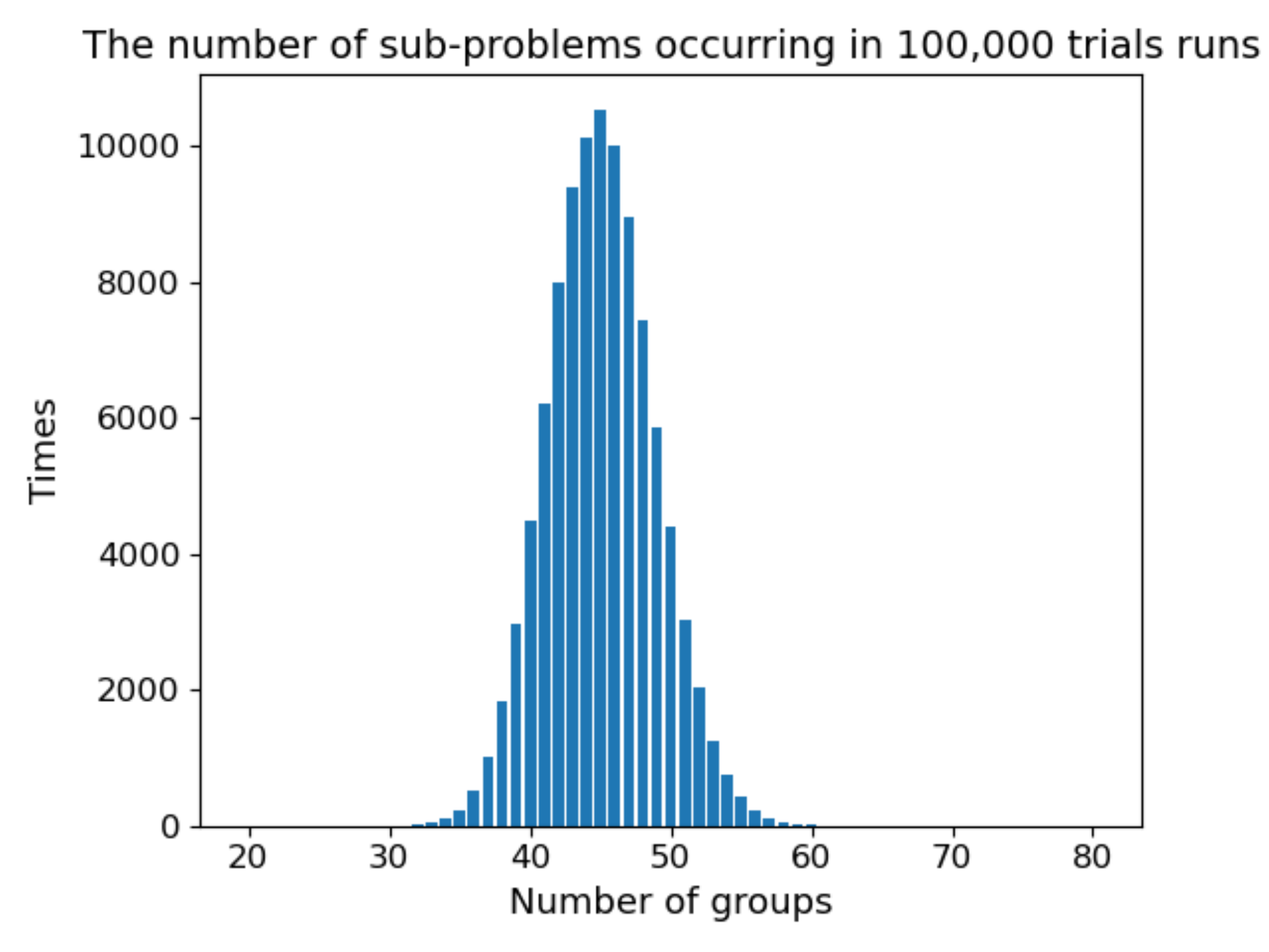}
	\caption{The number of sub-problems occurring in 100,000 trial runs}
	\label{fig:3}
\end{figure*}

The number of sub-problems occurring subjects to the normal distribution, and the most frequent occurrence is $45$. The average, minimum, and maximum size of sub-problems are also calculated, which are $22.86$, $1$, and $76$, respectively. Therefore, we can approximately consider that aRG decomposes the $1000$-D problems into $45 \times 23$-D ($1000$-D $\approx 45 \times 23$-D). 

\begin{figure*}[htb]
	\centering
	\includegraphics[width=12cm]{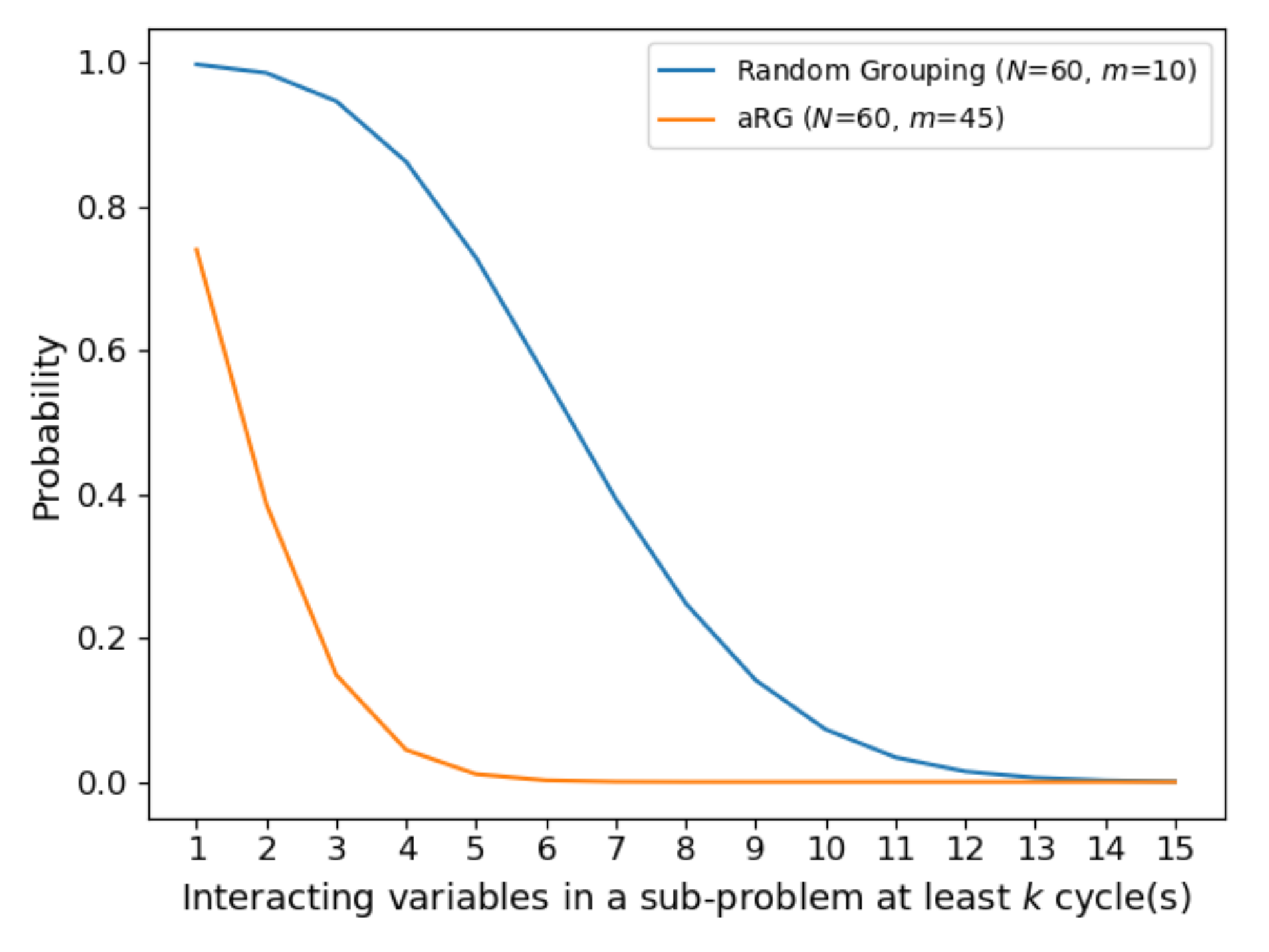}
	\caption{The probability of two interacting variables divided into a sub-problem at least $k$ cycle(s) in total $N$ times.}
	\label{fig:4}
\end{figure*}

Given that CEC2013 LSGO Suite has $3,000,000$ FEs limitation for $1000$-D problems, and the population size is set to $50$ for each sub-problems, thus the maximum generation $G=3,000,000/50/1000=60$. As aRG is applied in every generation of optimization, we can calculate that $P_1=0.7403, P_2=0.3862$ in 1 independent trial run based on Eq (\ref{eq:8}). Fig. \ref{fig:4} shows that the probability change with the increasing $k$ both on Random Grouping and aRG.

Although aRG is inferior to DECC-G in probabilistic analysis, the average size of sub-problems decomposed by aRG and DECC-G are $22.86$-D and $100$-D respectively. The smaller scale is easier for EAs to optimize especially under the FEs limitation. Notice that all the computation of aRG is an approximate estimation because of the high randomness of aRG. 

In summary, the whole Pseudocode of our proposal MDS-DSCC-aRG is shown in Algorithm \ref{alg:3}
\begin{algorithm}
	\label{alg:3}
	\DontPrintSemicolon
	\SetAlgoLined
	\KwIn {${\rm Dimension}:D; {\rm Population\ size}:s; {\rm Search\ space}:lb,ub; {\rm Max \ generation}:G$}
	\KwOut {${\rm Optimum}: x_{best}$}
	\SetKwFunction{FDE}{\textbf{MDS-DSCC-aRG}}
	\SetKwProg{Fn}{Function}{:}{}
	\Fn{\FDE{$D,s,lb,ub,G$}}{
		$CV \gets \textbf{zero}(D)$ \ // Context vector \;
		\While{$i=0$ \ \textbf{and} \ $i<G$}{
			$Gs \gets \textbf{aRG}(D)$ \ // Variable grouping in every generation\; 
			$t \gets \textbf{size}(Gs)$ \;
			\For{$j=0 \ to \ t$}{
				$x_{best, j} \gets \textbf{MDE-DS}(s, D, G, lb, ub, Gs_j, CV)$ \ // Context vector participates the evaluation\; 
				$CV \gets \textbf{update}(x_{best, j})$ \;
			}
		}
		$\textbf{return} \ x_{best}$
	}
	\caption{MDS-DSCC-aRG}
\end{algorithm}

\section{Numerical experiments and analysis} \label{sec:4}
In this section, we ran many experiments to measure our proposal, MDE-DSCC-aRG. In section \ref{sec:4.1}, we introduce the experimental settings, including benchmark functions, comparing methods, parameters of algorithms, and performance indicators. In section \ref{sec:4.2}, we provide the experiment results. Finally, we analyze our proposal both in the decomposition stage and optimization stage in section \ref{sec:4.3}.

\subsection{Experimental settings} \label{sec:4.1}

\subsubsection{Benchmark functions} \label{sec:4.1.1}
The benchmark functions we applied include two categories, $500$-D and $1000$-D, and the details of the original benchmark functions are shown in Table \ref{tbl:1}
\begin{table}[tbh]
	\scriptsize
	\centering
	\renewcommand\arraystretch{2}
	\caption{The benchmark functions of our experiment}
	\label{tbl:1}
	\begin{tabular}{ccccc}
		\toprule
		Benchmark & Dimensions & Equation & Search space &Separability \\
		\midrule
		Sphere & $500$-D & $f(x)=\sum_{i=1}^{D}x_i^2$ & $[-100, 100]$ & Separable	\\ 
		Rastrigin & $500$-D & $f(x)=10D+\sum_{i=1}^{D}(x_i^2-10\cos(2 \pi x_i))$ & $[-5.12, 5.12]$ & Separable \\ 
		Ackley & $500$-D & $f(x)=-20\exp(-0.2\sqrt{\frac{1}{D}\sum_{i=1}^{D}x_i^2})-\exp(\frac{1}{D}\sum_{i=1}^{D}\cos(2\pi x_i)) + 20 + e $ & $[-32.768, 32.768]$ & Separable	\\
		Rosenbrock & $500$-D & $f(x)=\sum_{i=1}^{D-1}(100(x_{i+1}-x_i^2)^2+(x_i-1)^2)$ & $[-30, 30]$ & Nonseparable	\\ 
		Dixon-Price & $500$-D & $f(x)=(x_1 -1)^2+\sum_{i=2}^{D}i(2x_i^2-x_{i-1})^2$ & $[-10, 10]$ & Nonseparable	\\
		CEC2013 LSGO Suite & $1000$-D & \multicolumn{3}{c}{referred in \cite{Li:13}}	\\ 
		\bottomrule
	\end{tabular}
\end{table}

There are 15 benchmark functions in CEC2013 LSGO Suite, so totally we have 20 functions to evaluate our proposal. Eq (\ref{eq:9}) shows our benchmark functions in multiplicative noisy environments. 
\begin{equation}
	\begin{aligned}
		\label{eq:9}
		f_i^{N}({\rm X}) = f_i({\rm X}) \cdot (1+\beta), i \in [1, 20]
	\end{aligned}
\end{equation}
And the Gaussian noise $\beta \sim N(0,0.01)$. 

\subsubsection{Comparing methods} \label{sec:4.1.2}
To measure the performance of our proposal in noisy environments, we compare aRG with some advanced methods, including Delta Grouping(D)\cite{Omidvar:10}, Differential Grouping (DG)\cite{Omidvar:14}, Random Grouping (G)\cite{Yang:08}, Multilevel Grouping (ML)\cite{Ke:08}, and Variable Interaction Learning(VIL)\cite{Chen:10}. We also compare the MDE-DS with the canonical DE to show the performance of MDE-DS in noisy environments. The compared algorithms are listed in Table \ref{tbl:2}
\begin{table}[tbh]
	\scriptsize
	\centering
	\renewcommand\arraystretch{1.3}
	\caption{A summary of the algorithms under comparison}
	\label{tbl:2}
	\begin{tabular}{ccc}
		\toprule
		Algorithms & Optimizer & Grouping methods \\
		\midrule
		DECC-D & \multirow{6}{*}{canonical DE} & D 	\\ 
		DECC-DG & ~ & DG 	\\ 
		DECC-G & ~ & G  \\ 
		DECC-ML & ~  & ML 	\\
		DECC-VIL & ~ & VIL 	\\
		DECC-aRG & ~ & \multirow{2}{*}{aRG} 	\\
		MDE-DSCC-aRG & MDE-DS & ~ \\ 
		\bottomrule
	\end{tabular}
\end{table}

\subsubsection{Parameters} \label{sec:4.1.3}
The explanation of parameters consists of two different stages: the decomposition stage and the optimization stage. Table \ref{tbl:3} shows the parameters of different grouping methods, Table \ref{tbl:4} shows the common parameters in the optimization stage. And the mutation strategy, scale factor, and crossover rate of canonical DE are DE/rand/1, $0.7$, and $0.9$ respectively. 

\begin{table}[tbh]
	\scriptsize
	\centering
	\renewcommand\arraystretch{1.3}
	\caption{The parameters of decomposition}
	\label{tbl:3}
	\begin{tabular}{ccc}
		\toprule
		Grouping methods & Parameters & Value \\
		\midrule
		\multirow{2}{*}{D} & sub-problem size &	$100$-D \\ 
		~ & number of sub-problems & 5 or 10 \\
		\midrule
		DG & $\varepsilon$ in nonlinearity check & $0.001$	\\ 
		\midrule
		\multirow{2}{*}{G} & sub-problem size &	$100$-D \\ 
		~ & number of sub-problems & 5 or 10 \\
		\midrule
		ML & number of sub-problem candidates  & [$5, 10, 25, 50$]	\\
		\bottomrule
	\end{tabular}
\end{table}

\begin{table}[tbh]
	\scriptsize
	\centering
	\renewcommand\arraystretch{1.3}
	\caption{The common parameters of sub-problems optimization}
	\label{tbl:4}
	\begin{tabular}{cc}
		\toprule
		Parameters & Value \\
		\midrule
		Optimization direction & Minimization \\
		trial runs & 25 \\
		FEs (grouping + optimization) & 750,000 or 3,000,000 \\
		Population size & 50 \\
		\bottomrule
	\end{tabular}
\end{table}

\subsubsection{Performance indicators} \label{sec:4.1.4}
Kruskal-Wallis test is applied to the fitness at the end of the optimization in 25 trial runs between different grouping methods, if the significance exists, then we apply the p-value acquired from the Mann-Whitney U test to do the Holm test. If our proposal is significantly better than the second-best algorithm, we mark \textcolor{red}{*} (significance level $5\%$) or \textcolor{red}{**} (significance level $10\%$) at the end of convergence. Meanwhile, we apply the Mann-Whitney U test between DECC-aRG and MDE-DSCC-aRG. If MDE-DSCC-aRG is significantly better than the DECC-aRG, then we mark \textcolor{orange}{*} or \textcolor{orange}{**} at the end of convergence.

\subsection{Experiment results} \label{sec:4.2}
In this section, the performance of our proposal is studied. Experiments are conducted on the benchmark functions presented in section \ref{sec:4.1.1}. Table \ref{tbl:5} shows the mean and standard deviation of 25 independent trial runs within DECC-D, DECC-G, DECC-ML, DECC-DG, DECC-VIL, and DECC-aRG to verify the performance of aRG. Table \ref{tbl:6} compares the DECC-aRG and MDE-DSCC-aRG to show the efficiency of the introduction of MDE-DS. The best solution is in bold. 

\begin{sidewaystable}[tbh]
	\scriptsize
	\centering
	\caption{The mean and std of DECC-D, DECC-G, DECC-ML, DECC-DG, DECC-VIL, and DECC-aRG in 25 trial runs}
	\label{tbl:5}
	\begin{tabular}{ccccccccccccc}
		\toprule
		\multirow{2}{*}{Func.} & \multicolumn{2}{c}{DECC-D} & \multicolumn{2}{c}{DECC-G} &  \multicolumn{2}{c}{DECC-ML} & \multicolumn{2}{c}{DECC-DG} & \multicolumn{2}{c}{DECC-VIL} & \multicolumn{2}{c}{DECC-aRG} \\
		\cmidrule(r){2-3} \cmidrule(r){4-5} \cmidrule(r){6-7} \cmidrule(r){8-9} \cmidrule(r){10-11} \cmidrule(r){12-13}
		~ & mean & std & mean & std & mean & std & mean & std & mean & std & mean & std  \\
		\midrule 
		$f_1$ & 5.97e+07 & 3.89e+06 & 5.65e+07 & 5.52e+06 & 4.63e+07 & 1.16e+07 & 6.54e+08 & 2.61e+07 & 5.19e+08 & 9.49e+07 &  \textbf{2.04e+06} &  \textbf{1.41e+05}  \\
		$f_2$ & 1.72e+04 & 2.73e+02 & 1.71e+04 & 2.34e+02 & 1.68e+04 & 3.01e+02 & 2.47e+04 & 5.01e+02 & 2.34e+04 & 7.38e+02 & \textbf{1.51e+04} & \textbf{1.69e+02}  \\
		$f_3$ & 2.16e+01 & 8.37e-03 & 2.16e+01 & 7.06e-03 & \textbf{2.16e+01} & \textbf{9.32e-03} & 2.16e+01 & 8.09e-03 & 2.16e+01 & 9.15e-03 & 2.16e+01 & 6.11e-03  \\
		$f_4$ & 7.92e+11 & 3.58e+11 & 6.99e+11 & 4.35e+11 & 5.43e+11 & 3.81e+11 & 7.02e+11 & 2.27e+11 & 8.04e+11 & 3.33e+11 & \textbf{3.87e+11} & \textbf{1.00e+11} \\
		$f_5$ & 1.37e+07 & 1.07e+06 & 9.18e+06 & 5.76e+06 & \textbf{3.54e+06} & \textbf{6.40e+05} & 1.35e+07 & 1.41e+06 & 1.35e+07 & 1.56e+06 & 7.71e+06 & 1.03e+06 \\
		$f_6$ & 1.06e+06 & 1.24e+03 & 1.06e+06 & 1.76e+03 & \textbf{1.06e+06} & \textbf{1.17e+03} & 1.06e+06 & 1.11e+03 & 1.06e+06 & 1.06e+03 & 1.06e+06 & 1.30e+03 \\
		$f_7$ & 4.40e+09 & 2.23e+09 & 3.76e+09 & 2.49e+09 & \textbf{2.76e+09} & \textbf{1.60e+09} & 3.59e+09 & 1.51e+09 & 3.35e+09 & 1.54e+09 & 2.83e+09 & 1.05e+09  \\
		$f_8$ & 3.42e+16 & 1.89e+16 & 1.57e+16 & 1.39e+16 & 2.28e+16 & 1.58e+16 & 4.39e+16 & 1.76e+16 & 2.47e+16 & 1.18e+16 & \textbf{1.28e+16} & \textbf{3.26e+15}  \\
		$f_9$ & 1.01e+09 & 7.07e+07 & 9.86e+08 & 8.45e+07 & \textbf{3.24e+08} & \textbf{8.67e+07} & 9.91e+08 & 9.39e+07 & 1.01e+09 & 1.04e+08 & 7.07e+08 & 1.08e+08  \\
		$f_{10}$ & 9.55e+07 & 2.30e+05 & 9.55e+07 & 2.75e+05 & 9.55e+07 & 3.24e+05 & 9.53e+07 & 3.47e+05 & 9.54e+07 & 3.66e+05 & \textbf{9.53e+07} & \textbf{2.59e+05} \\
		$f_{11}$ & 5.32e+11 & 2.54e+11 & 4.14e+11 & 2.44e+11 & \textbf{3.13e+11} & \textbf{1.60e+11} & 3.71e+11 & 1.54e+11 & 4.02e+11 & 2.32e+11 & 5.53e+11 & 2.51e+11 \\
		$f_{12}$ & 1.91e+06 & 3.51e+05 & 1.82e+06 & 5.80e+05 & 1.24e+06 & 9.13e+05 & 1.97e+09 & 2.30e+08 & 1.47e+09 & 3.52e+08 & \textbf{1.04e+04} & \textbf{7.05e+02}  \\
		$f_{13}$ & 6.12e+10 & 1.26e+10 & 4.39e+10 & 1.99e+10 & 2.18e+10 & 5.22e+09 & 5.39e+10 & 1.85e+10 & 5.34e+10 & 1.77e+10 & \textbf{2.08e+10} & \textbf{3.38e+09} \\
		$f_{14}$ & 6.88e+11 & 3.22e+11 & 4.80e+11 & 2.79e+11 & \textbf{3.74e+11} & \textbf{1.92e+11} & 5.67e+11 & 1.46e+11 & 8.37e+11 & 3.34e+11 & 4.08e+11 & 1.57e+11 \\
		$f_{15}$ & 2.28e+08 & 3.38e+07 & 2.39e+08 & 3.48e+07 & 1.55e+08 & 1.51e+07 & 1.57e+08 & 3.75e+07 & 1.63e+08 & 2.23e+07 & \textbf{4.72e+07} & \textbf{3.85e+06}  \\
		Sphere & 1.54e+04 & 1.24e+03 & 1.42e+04 & 1.19e+03 & 1.22e+03 & 4.83e+02 & 4.11e+04 & 3.30e+03 & 6.32e+01 & 1.16e+01 & \textbf{6.19e+01} & \textbf{8.92e+00} \\
		Rastrigin & 7.18e+03 & 1.12e+02 & 7.18e+03 & 1.23e+02 & 7.10e+03 & 1.38e+02 & 7.46e+03 & 1.04e+02 & \textbf{6.47e+03} & \textbf{8.64e+01} & 7.21e+03 & 9.81e+01 \\
		Ackley & 2.06e+01 & 4.95e-02 & 2.06e+01 & 6.85e-02 & 2.06e+01 & 7.99e-02 & 2.05e+01 & 9.28e-02 & \textbf{1.98e+01} & \textbf{1.03e-01} & 2.06e+01 & 8.26e-02 \\
		Rosenbrock & 2.80e+05 & 2.94e+04 & 2.21e+05 & 3.99e+04 & 1.86e+04 & 8.53e+03 & 2.44e+06 & 4.71e+05 & \textbf{2.93e+03} & \textbf{2.70e+02} & 4.76e+03 & 3.99e+02 \\
		Dixon-Price	& 4.07e+04 & 4.31e+03 & 3.91e+04 & 6.37e+03 & 4.81e+03 & 1.14e+03 & 2.14e+05 & 3.28e+04 & 4.36e+04 & 2.01e+04 & \textbf{1.37e+03} & \textbf{2.19e+02} \\ 
		\bottomrule
	\end{tabular}
\end{sidewaystable}

\begin{figure*}[htb]
	\centering
	\includegraphics[width=15cm]{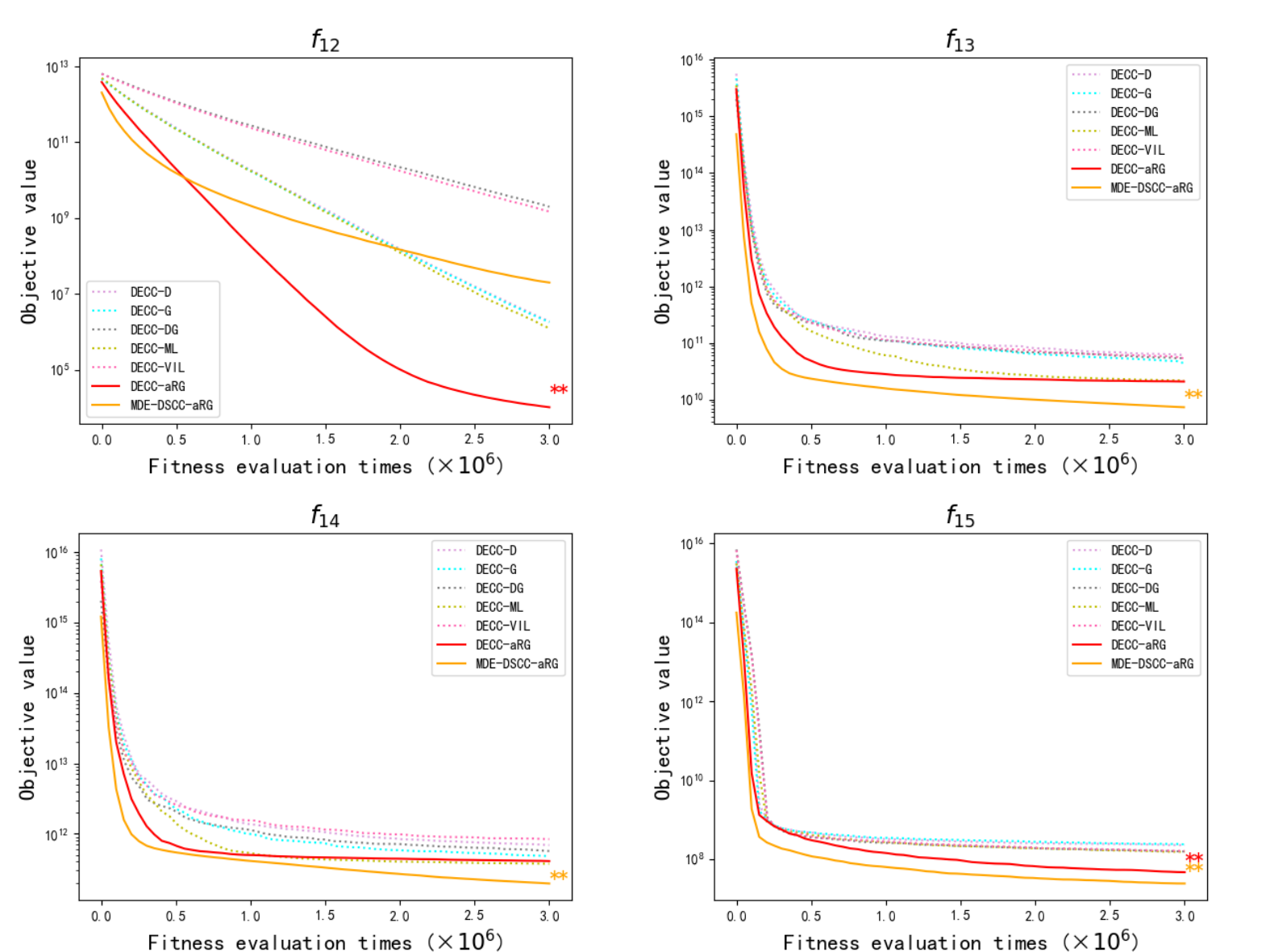}
	\caption{The convergence curve of DECC-D, DECC-G, DECC-DG, DECC-ML, DECC-VIL, and DECC-aRG in 25 trial runs}
	\label{fig:5}
\end{figure*}

\begin{table}[tbh]
	\scriptsize
	\centering
	\caption{The mean and std of DECC-aRG and MDE-DSCC-aRG in 25 trial runs}
	\label{tbl:6}
	\begin{tabular}{ccccc}
		\toprule
		\multirow{2}{*}{Func.} & \multicolumn{2}{c}{DECC-aRG} & \multicolumn{2}{c}{MDE-DSCC-aRG} \\
		\cmidrule(r){2-3} \cmidrule(r){4-5} 
		~ & mean & std & mean & std \\
		\midrule 
		$f_1$ & 2.04e+06 & 1.41e+05 & \textbf{9.84e+03} & \textbf{1.17e+04} \\
		$f_2$ & 1.51e+04 & 1.69e+02 & \textbf{2.52e+03} & \textbf{5.80e+01} \\
		$f_3$ & \textbf{2.16e+01} & \textbf{6.11e-03} & 2.16e+01 & 8.53e-03 \\
		$f_4$ & 3.87e+11 & 1.00e+11 & \textbf{1.15e+11} & \textbf{5.39e+10} \\
		$f_5$ & \textbf{7.71e+06} & \textbf{1.03e+06} & 9.16e+06 & 1.30e+06 \\
		$f_6$ & \textbf{1.06e+06} & \textbf{1.30e+03} & 1.06e+06 & 1.31e+03 \\
		$f_7$ & 2.83e+09 & 1.05e+09 & \textbf{7.74e+08} & \textbf{3.16e+08} \\
		$f_8$ & 1.28e+16 & 3.26e+15 & \textbf{5.00e+15} & \textbf{2.00e+15} \\
		$f_9$ & \textbf{7.07e+08} & \textbf{1.08e+08} & 7.88e+08 & 1.80e+08 \\
		$f_{10}$ & 9.53e+07 & 2.59e+05 & \textbf{9.48e+07} & \textbf{9.93e+05} \\
		$f_{11}$ & 5.53e+11 & 2.51e+11 & \textbf{1.83e+11} & \textbf{1.22e+11} \\
		$f_{12}$ & \textbf{1.04e+04} & \textbf{7.05e+02} & 2.00e+07 & 1.81e+07 \\
		$f_{13}$ & 2.08e+10 & 3.38e+09 & \textbf{7.34e+09} & \textbf{2.75e+09} \\
		$f_{14}$ & 4.08e+11 & 1.57e+11 & \textbf{1.95e+11} & \textbf{6.06e+10} \\
		$f_{15}$ & 4.72e+07 & 3.85e+06 & \textbf{2.44e+07} & \textbf{2.17e+06} \\
		Sphere & \textbf{6.19e+01} & \textbf{8.92e+00} & 3.20e+02 & 1.94e+01 \\
		Rosenbrock & 4.76e+03 & 3.99e+02 & \textbf{3.75e+03} & \textbf{6.20e+02} \\
		Ackley & 2.06e+01 & 8.26e-02 & \textbf{2.00e+01} & \textbf{4.24e-02} \\
		Dixon-Price & 1.37e+03 & 2.19e+02 & \textbf{1.02e+03} & \textbf{2.12e+02} \\
		Rastrigin & 7.21e+03 & 9.81e+01 & \textbf{5.16e+03} & \textbf{7.01e+01} \\
		\bottomrule
	\end{tabular}
\end{table}
Meanwhile, the convergence curve of compared methods is provided in Fig. \ref{fig:5}.

\begin{figure*}[htb]
	\centering
	\includegraphics[width=15cm]{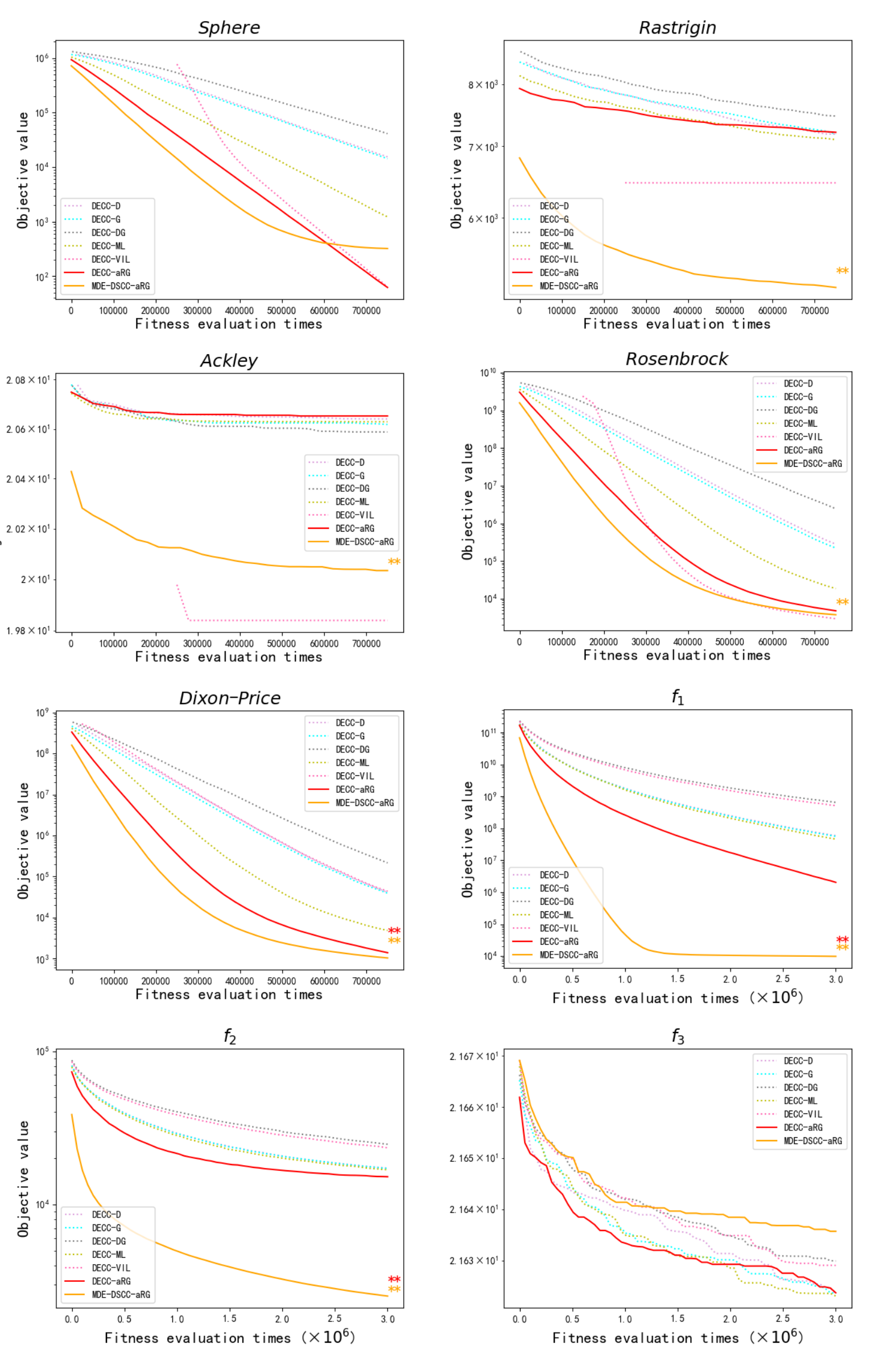}
\end{figure*}
\begin{figure*}[htb]
	\centering
	\includegraphics[width=15cm]{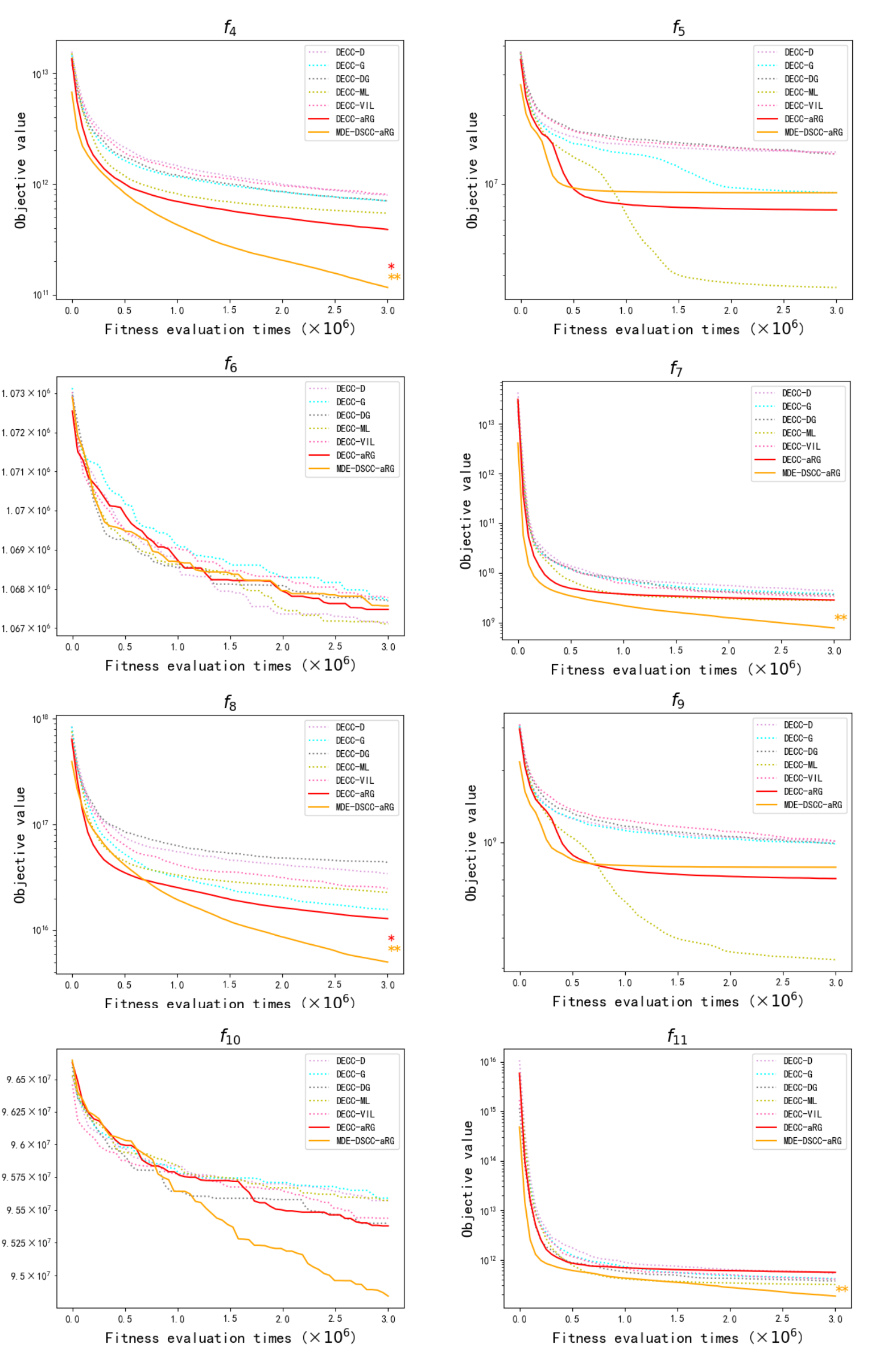}
\end{figure*}

\subsection{Analysis} \label{sec:4.3}
In this section, we analyze the performance of our proposal in both the decomposition method of aRG and the introduction of MDE-DS. 

\subsubsection{aRG vs. DG}  \label{sec:4.3.1}
As we state the challenge of nonlinearity check-based methods in noisy environments in section \ref{sec:2.3}, DG cannot detect the interactions between variables in noisy environments in practice. In DECC-DG, all variables are assigned to a group and optimized directly, and it is difficult for canonical DE to find an acceptable solution in large-scale problems. Thus, our proposed aRG performs better than DG in noisy environments, and DG is the most environmentally sensitive grouping method among the compared methods. 

\subsubsection{aRG vs. VIL}  \label{sec:4.3.2}
VIL applies the Eq (\ref{eq:10}) to identify the interaction between $x_i$ and $x_j$: 
\begin{equation}
	\begin{aligned}
		\label{eq:10}
		if \ \exists s,s_i,s_j,s_{ij}:\ f(s) > f(s_i)  \ and \ f(s_j) < f(s_{ij}) \\
		then \  x_{i} \  and \  x_{j} \  are \  nonseparable
	\end{aligned}
\end{equation}
Monotonicity detection is employed in VIL. When $x_{i}$ and $x_{j}$ satisfy the simultaneous increase or decrease within finite samples, VIL identifies $x_{i}$ and $x_{j}$ as separable variables. In theory, VIL has the probability to detect interactions even in noisy environments. In $500$-D problems without shift or rotation process in noisy environments, experimental results show that VIL can identify some interactions and achieve good performance. But in the complex fitness landscape with noise, only a few interactions can be identified and the performance of VIL is close to DG. From the above analysis, aRG performs better than VIL in complex noisy environments and VIL is the second most environmentally sensitive grouping method among the compared methods.

\subsubsection{aRG vs. D}  \label{sec:4.3.3}
\begin{figure}[htb]
	\centering
	\includegraphics[width=15cm]{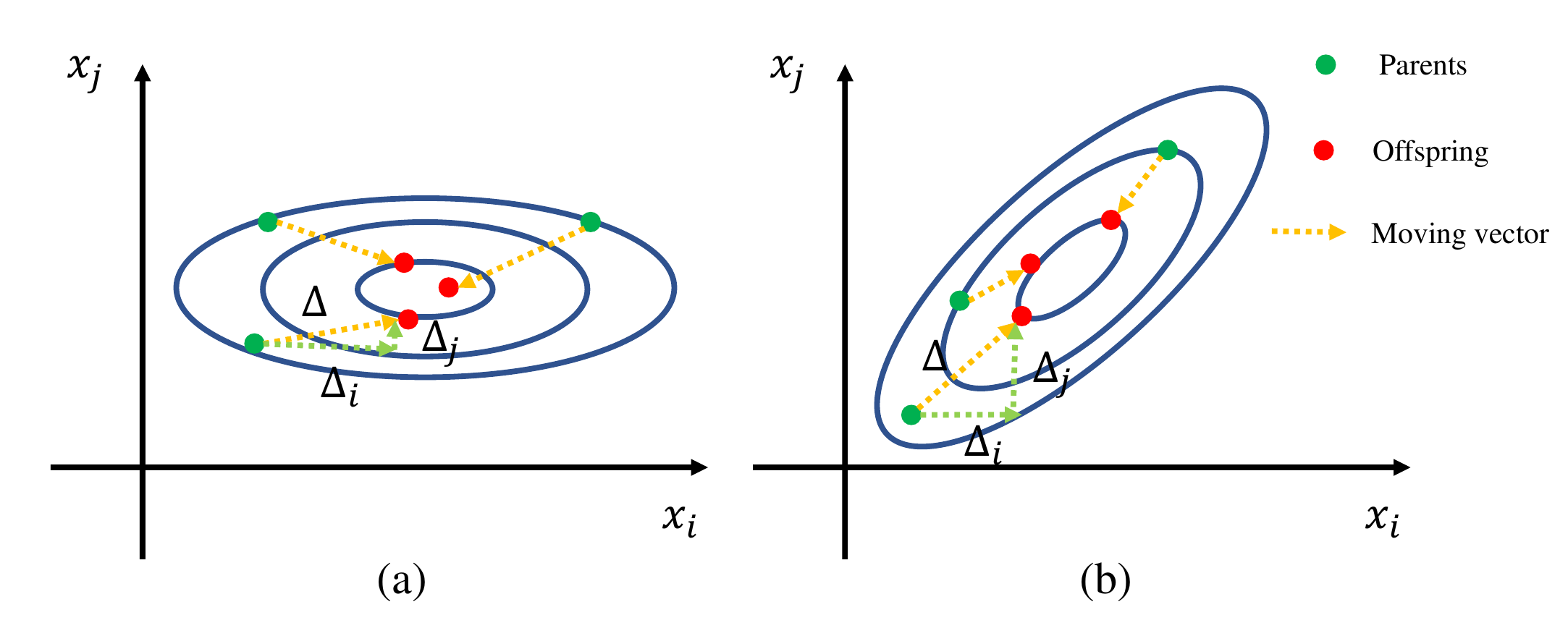}
	\caption{(a). Delta Grouping works on the separable function. (b). Delta Grouping works on the nonseparable function.}
	\label{fig:6}
\end{figure}

The schematic diagram of Delta Grouping is shown in Fig \ref{fig:6}.

Delta Grouping realizes that the difference in coordinates from the initial random population to the optimized population is different in the separable function and the nonseparable function. In Fig \ref{fig:6}, when the $\Delta_{i}$ and $\Delta_{j}$ has large difference, Delta Grouping identify $x_{i}$ and $x_{j}$ are separable. This rough estimation is still affected by the noise because the Delta Grouping samples in the fitness landscape, and $m, k$ specified by the user limit the Delta Grouping.

\subsubsection{aRG vs. Random Grouping}  \label{sec:4.3.4}
As we analyze the difference between aRG and Random Grouping in section \ref{sec:3}, aRG has a lower probability to detect the interactions but the scale of sub-problems in aRG is also smaller. Thus, aRG is better than Random Grouping in separable functions in theory, and experiment results prove that. In partially separable and nonseparable functions, the competitiveness of ML increases as the interactions between variables increase. However, the scale of sub-problems still limits the DECC-G and DECC-ML to achieve better solutions. Therefore, aRG is competitive with Random Grouping in partially separable and nonseparable functions. Without the requirement of fitness landscape information, the Random Grouping methods are the most environmentally insensitive. 

\subsubsection{The efficiency of MDE-DS}  \label{sec:4.3.5}
Figure \ref{fig:5} and Table \ref{tbl:6} all prove that MDE-DS has strong ability to search better solutions comparing with the canonical DE in most benchmark functions, although canonical DE performs better in $f_3$, $f_5$, $f_6$,  $f_9$,  $f_{12}$, and Sphere functions. However, no optimization algorithm can solve all optimization problems perfectly. According to No Free Lunch Theory\cite{Wolpert:97} in optimization, the average performance of any pair of algorithms A and B is identical on all possible problems. Therefore, if an algorithm performs well on a certain class of problems, it must pay for that with performance degradation on the remaining problems, since this is the only way for all algorithms to have the same performance on average across all functions. Thus, although MDE-DS may perform worse than the canonical DE in noiseless functions, it is successful to introduce MDE-DS to solve problems in noisy environments.  

\section{Discussion} \label{sec:5}
The above analyses show our proposal has broad prospects to solve large-scale optimization problems in noisy environments. However, there are still many aspects for improvement. Here, we list some open topics for potential and future research.

\subsection{Interactions identification in noisy environments}
As we describe in section \ref{sec:3}, randomness can capture the interactions between variables to some extent, however, Figure \ref{fig:4} shows that this probability will degrade dramatically as $k$ increases. Therefore, it is necessary to develop an efficient interaction identification method in noisy environments. Explicit averaging\cite{Laura:95} can alleviate the uncertainty of noise by re-evaluation. Let the re-evaluating times for $f^N({\rm X})$ be $m$ and $f^N_i({\rm X})$) represents the $i^{th}$ re-evaluation value. Then we apply the principle of Monte Carlo integration\cite{Gayathri:01}, the mean fitness estimation $f^N({\rm X})$, standard deviation $\sigma (f^N({\rm X}))$ and the standard error of the mean fitness $se(f^N({\rm X}))$ are calculated as
\begin{equation}
	\begin{aligned}
		\label{eq:11}
		\bar{f}^N({\rm X})=\frac{1}{m}\sum_{i=1}^{m}f_i^N({\rm X}) \\
		\sigma ({f}^N({\rm X}))=\sqrt{\frac{1}{m-1}\sum_{i=1}^{m}(f_i^N({\rm X})-\bar{f}^N({\rm X}))} \\
		se(\bar{f}^N({\rm X}))=\frac{\sigma ({f}^N({\rm X}))}{\sqrt{m}}
	\end{aligned}
\end{equation}

Eq (\ref{eq:11}) shows that sampling an individualfs objective function $m$ times can reduce  $se(\bar{f}^N({\rm X}))$ by a factor of $m$ to improve the accuracy in the mean fitness estimation, which means the accuracy of sampling increases. It is a feasible method to loosen the threshold $\varepsilon$ in DG and combine the explicit averaging strategy to identify the interactions, although it will consume lots of FEs. 

\subsection{The balance between the scale and the number of sub-problems}
Due to the curse of dimensionality, the performance of optimization algorithms will degenerate extremely as the problem scale rises in a certain dimensional range. In future research, our research direction includes two aspects, (1). Improving the ability of the optimization algorithm to solve higher dimensional problems. (2). Finding the optimal problem scale corresponding to a certain optimization algorithm through pre-experiment, reinforcement learning, and other methods. During the variable grouping process,  the grouping algorithm will reject adding more variables to the sub-problem when the size of the sub-problem reaches the optimal scale.

\section{Conclusion} \label{sec:6}
In this paper, we propose an automatic random grouping and introduce MDE-DS to enhance the search ability in noisy environments. Numerical experiments show that our proposal is competitive with the compared grouping methods in noisy environments.

In future research, we will focus on the development of efficient interaction identification methods in noisy environments and the balance of the number and the scale of sub-problems.

\bibliographystyle{elsarticle-num}
\bibliography{IS}

\begin{thebibliography}{10}
\expandafter\ifx\csname url\endcsname\relax
  \def\url#1{\texttt{#1}}\fi
\expandafter\ifx\csname urlprefix\endcsname\relax\def\urlprefix{URL }\fi
\expandafter\ifx\csname href\endcsname\relax
  \def\href#1#2{#2} \def\path#1{#1}\fi

\bibitem{Yang:08}
Z.~Yang, K.~Tang, X.~Yao, Large scale evolutionary optimization using
  cooperative coevolution, Information Sciences 178~(15) (2008) 2985--2999,
  nature Inspired Problem-Solving.
\newblock \href {http://dx.doi.org/https://doi.org/10.1016/j.ins.2008.02.017}
  {\path{doi:https://doi.org/10.1016/j.ins.2008.02.017}}.

\bibitem{Thomas:93}
T.~Back, H.-P. Schwefel, An overview of evolutionary algorithms for parameter
  optimization, Evolutionary Computation 1~(1) (1993) 1--23.
\newblock \href {http://dx.doi.org/10.1162/evco.1993.1.1.1}
  {\path{doi:10.1162/evco.1993.1.1.1}}.

\bibitem{Darrell:01}
D.~Whitley, An overview of evolutionary algorithms: practical issues and common
  pitfalls, Information and Software Technology 43~(14) (2001) 817--831.
\newblock \href
  {http://dx.doi.org/https://doi.org/10.1016/S0950-5849(01)00188-4}
  {\path{doi:https://doi.org/10.1016/S0950-5849(01)00188-4}}.

\bibitem{Mario:00}
M.~K{\"o}ppen, The curse of dimensionality, in: 5th online world conference on
  soft computing in industrial applications (WSC5), Vol.~1, 2000, pp. 4--8.

\bibitem{Martinsson:20}
P.-G. Martinsson, J.~A. Tropp, Randomized numerical linear algebra: Foundations
  and algorithms, Acta Numerica 29 (2020) 403--572.
\newblock \href {http://dx.doi.org/doi:10.1017/S0962492920000021}
  {\path{doi:doi:10.1017/S0962492920000021}}.

\bibitem{Bottou:18}
L.~Bottou, F.~E. Curtis, J.~Nocedal, Optimization methods for large-scale
  machine learning, SIAM Review 60~(2) (2018) 223--311.
\newblock \href {http://dx.doi.org/10.1137/16M1080173}
  {\path{doi:10.1137/16M1080173}}.

\bibitem{Gould:05}
N.~Gould, D.~Orban, P.~Toint^^c2^^a0, Numerical methods for large-scale
  nonlinear optimization, Acta Numerica 14 (2005) 299 -- 361.
\newblock \href {http://dx.doi.org/10.1017/S0962492904000248}
  {\path{doi:10.1017/S0962492904000248}}.

\bibitem{Laura:95}
L.~Painton, U.~Diwekar, Stochastic annealing for synthesis under uncertainty,
  European Journal of Operational Research 83~(3) (1995) 489--502.
\newblock \href
  {http://dx.doi.org/https://doi.org/10.1016/0377-2217(94)00245-8}
  {\path{doi:https://doi.org/10.1016/0377-2217(94)00245-8}}.

\bibitem{Diaz:15}
J.~Diaz, J.~Handl, Implicit and explicit averaging strategies for
  simulation-based optimization of a real-world production planning problem,
  Informatica (Slovenia) 39 (2015) 161--168.

\bibitem{Youhei:15}
Y.~Akimoto, S.~Astete-Morales, O.~Teytaud, Analysis of runtime of optimization
  algorithms for noisy functions over discrete codomains, Theoretical Computer
  Science 605 (2015) 42--50.
\newblock \href {http://dx.doi.org/https://doi.org/10.1016/j.tcs.2015.04.008}
  {\path{doi:https://doi.org/10.1016/j.tcs.2015.04.008}}.

\bibitem{Chen:20}
Y.-W. Chen, Q.~Song, X.~Liu, P.~S. Sastry, X.~Hu, On robustness of neural
  architecture search under label noise, Frontiers in Big Data 3.
\newblock \href {http://dx.doi.org/10.3389/fdata.2020.00002}
  {\path{doi:10.3389/fdata.2020.00002}}.

\bibitem{Qian:17}
C.~Qian, J.-C. Shi, Y.~Yu, K.~Tang, Z.-H. Zhou, Subset selection under noise,
  Advances in neural information processing systems 30.

\bibitem{Potter:94}
M.~Potter, K.~De~Jong, A cooperative coevolutionary approach to function
  optimization, Lecture Notes in Computer Science (including subseries Lecture
  Notes in Artificial Intelligence and Lecture Notes in Bioinformatics) 866
  LNCS (1994) 249--257.

\bibitem{Sun:17}
Y.~Sun, M.~Kirley, S.~K. Halgamuge, A recursive decomposition method for large
  scale continuous optimization, IEEE Transactions on Evolutionary Computation
  22~(5) (2017) 647--661.
\newblock \href {http://dx.doi.org/10.1109/TEVC.2017.2778089}
  {\path{doi:10.1109/TEVC.2017.2778089}}.

\bibitem{Ma:16}
X.~Ma, F.~Liu, Y.~Qi, X.~Wang, L.~Li, L.~Jiao, M.~Yin, M.~Gong, A
  multiobjective evolutionary algorithm based on decision variable analyses for
  multiobjective optimization problems with large-scale variables, IEEE
  Transactions on Evolutionary Computation 20~(2) (2016) 275--298.
\newblock \href {http://dx.doi.org/10.1109/TEVC.2015.2455812}
  {\path{doi:10.1109/TEVC.2015.2455812}}.

\bibitem{Sayed:15}
E.~Sayed, D.~Essam, R.~Sarker, S.~Elsayed, Decomposition-based evolutionary
  algorithm for large scale constrained problems, Information Sciences 316
  (2015) 457--486, nature-Inspired Algorithms for Large Scale Global
  Optimization.
\newblock \href {http://dx.doi.org/https://doi.org/10.1016/j.ins.2014.10.035}
  {\path{doi:https://doi.org/10.1016/j.ins.2014.10.035}}.

\bibitem{Mei:14}
Y.~Mei, X.~Li, X.~Yao, Cooperative coevolution with route distance grouping for
  large-scale capacitated arc routing problems, IEEE Transactions on
  Evolutionary Computation 18~(3) (2014) 435--449.
\newblock \href {http://dx.doi.org/10.1109/TEVC.2013.2281503}
  {\path{doi:10.1109/TEVC.2013.2281503}}.

\bibitem{Tezuka:04}
M.~Tezuka, M.~Munetomo, K.~Akama, Linkage identification by nonlinearity check
  for real-coded genetic algorithms, Lecture Notes in Computer Science
  (including subseries Lecture Notes in Artificial Intelligence and Lecture
  Notes in Bioinformatics) 3103 (2004) 222--233.

\bibitem{Omidvar:14}
M.~N. Omidvar, X.~Li, Y.~Mei, X.~Yao, Cooperative co-evolution with
  differential grouping for large scale optimization, IEEE Transactions on
  Evolutionary Computation 18~(3) (2014) 378--393.
\newblock \href {http://dx.doi.org/10.1109/TEVC.2013.2281543}
  {\path{doi:10.1109/TEVC.2013.2281543}}.

\bibitem{Omidvar:17}
M.~N. Omidvar, M.~Yang, Y.~Mei, X.~Li, X.~Yao, D{G}2: {A} {F}aster and {M}ore
  {A}ccurate {D}ifferential {G}rouping for {L}arge-{S}cale {B}lack-{B}ox
  {O}ptimization, IEEE Transactions on Evolutionary Computation 21~(6) (2017)
  929--942.
\newblock \href {http://dx.doi.org/10.1109/TEVC.2017.2694221}
  {\path{doi:10.1109/TEVC.2017.2694221}}.

\bibitem{Sun:15}
Y.~Sun, M.~Kirley, S.~K. Halgamuge, Extended differential grouping for large
  scale global optimization with direct and indirect variable interactions, in:
  Proceedings of the 2015 Annual Conference on Genetic and Evolutionary
  Computation, GECCO '15, Association for Computing Machinery, New York, NY,
  USA, 2015, p. 313^^e2^^80^^93320.
\newblock \href {http://dx.doi.org/10.1145/2739480.2754666}
  {\path{doi:10.1145/2739480.2754666}}.

\bibitem{Ling:16}
Y.~Ling, H.~Li, B.~Cao, Cooperative co-evolution with graph-based differential
  grouping for large scale global optimization, in: 2016 12th International
  Conference on Natural Computation, Fuzzy Systems and Knowledge Discovery
  (ICNC-FSKD), 2016, pp. 95--102.
\newblock \href {http://dx.doi.org/10.1109/FSKD.2016.7603157}
  {\path{doi:10.1109/FSKD.2016.7603157}}.

\bibitem{Wu:22}
Y.~Wu, X.~Peng, H.~Wang, Y.~Jin, D.~Xu, Cooperative {C}oevolutionary {CMA-ES}
  with {L}andscape-{A}ware {G}rouping in {N}oisy {E}nvironments, IEEE
  Transactions on Evolutionary Computation (2022) 1--1\href
  {http://dx.doi.org/10.1109/TEVC.2022.3180224}
  {\path{doi:10.1109/TEVC.2022.3180224}}.

\bibitem{Ghosh:17}
A.~Ghosh, S.~Das, R.~Mallipeddi, A.~K. Das, S.~S. Dash, A modified differential
  evolution with distance-based selection for continuous optimization in
  presence of noise, IEEE Access 5 (2017) 26944--26964.
\newblock \href {http://dx.doi.org/10.1109/ACCESS.2017.2773825}
  {\path{doi:10.1109/ACCESS.2017.2773825}}.

\bibitem{Goldberg:96}
H.~Kargupta, D.~E. Goldberg, Blackbox optimization: Implications of search, in:
  In communication. Submitted in International Journal of Foundation of
  Computer Science, 1996, pp. 96--63.

\bibitem{Munetomo:99}
M.~Munetomo, D.~E. Goldberg, Linkage identification by non-monotonicity
  detection for overlapping functions, Evolutionary Computation 7~(4) (1999)
  377--398.
\newblock \href {http://dx.doi.org/10.1162/evco.1999.7.4.377}
  {\path{doi:10.1162/evco.1999.7.4.377}}.

\bibitem{Chen:10}
W.~Chen, T.~Weise, Z.~Yang, K.~Tang, Large-scale global optimization using
  cooperative coevolution with variable interaction learning, in: International
  Conference on Parallel Problem Solving from Nature, Springer, 2010, pp.
  300--309.

\bibitem{Singh:10}
H.~K. Singh, T.~Ray, Divide and Conquer in Coevolution: A Difficult Balancing
  Act, Springer Berlin Heidelberg, Berlin, Heidelberg, 2010, pp. 117--138.
\newblock \href {http://dx.doi.org/10.1007/978-3-642-13425-8\_6}
  {\path{doi:10.1007/978-3-642-13425-8\_6}}.

\bibitem{Rojas:11}
Y.~Rojas, R.~Landa, Towards the use of statistical information and differential
  evolution for large scale global optimization, in: 2011 8th International
  Conference on Electrical Engineering, Computing Science and Automatic
  Control, 2011, pp. 1--6.
\newblock \href {http://dx.doi.org/10.1109/ICEEE.2011.6106645}
  {\path{doi:10.1109/ICEEE.2011.6106645}}.

\bibitem{Ke:08}
Z.~Yang, K.~Tang, X.~Yao, Multilevel cooperative coevolution for large scale
  optimization, in: 2008 IEEE Congress on Evolutionary Computation (IEEE World
  Congress on Computational Intelligence), 2008, pp. 1663--1670.
\newblock \href {http://dx.doi.org/10.1109/CEC.2008.4631014}
  {\path{doi:10.1109/CEC.2008.4631014}}.

\bibitem{Song:17}
A.~Song, W.-N. Chen, P.-T. Luo, Y.-J. Gong, J.~Zhang, Overlapped cooperative
  co-evolution for large scale optimization, in: 2017 IEEE International
  Conference on Systems, Man, and Cybernetics (SMC), 2017, pp. 3689--3694.
\newblock \href {http://dx.doi.org/10.1109/SMC.2017.8123206}
  {\path{doi:10.1109/SMC.2017.8123206}}.

\bibitem{Omidvar:10}
M.~Omidvar, X.~Li, X.~Yao, Cooperative co-evolution with delta grouping for
  large scale non-separable function optimization, 2010, pp. 1--8.
\newblock \href {http://dx.doi.org/10.1109/CEC.2010.5585979}
  {\path{doi:10.1109/CEC.2010.5585979}}.

\bibitem{Dai:16}
G.~Dai, X.~Chen, L.~Chen, M.~Wang, L.~Peng, Cooperative coevolution with
  dependency identification grouping for large scale global optimization, in:
  2016 IEEE Congress on Evolutionary Computation (CEC), 2016, pp. 5201--5208.
\newblock \href {http://dx.doi.org/10.1109/CEC.2016.7748349}
  {\path{doi:10.1109/CEC.2016.7748349}}.

\bibitem{Storn:96}
R.~Storn, On the usage of differential evolution for function optimization, in:
  Proceedings of North American Fuzzy Information Processing, 1996, pp.
  519--523.
\newblock \href {http://dx.doi.org/10.1109/NAFIPS.1996.534789}
  {\path{doi:10.1109/NAFIPS.1996.534789}}.

\bibitem{He:09}
X.~He, Q.~Zhang, N.~Sun, Y.~Dong, Feature selection with discrete binary
  differential evolution, in: 2009 International Conference on Artificial
  Intelligence and Computational Intelligence, Vol.~4, 2009, pp. 327--330.
\newblock \href {http://dx.doi.org/10.1109/AICI.2009.438}
  {\path{doi:10.1109/AICI.2009.438}}.

\bibitem{Du:07}
J.-X. Du, D.-S. Huang, X.-F. Wang, X.~Gu, Shape recognition based on neural
  networks trained by differential evolution algorithm, Neurocomputing 70~(4)
  (2007) 896--903, advanced Neurocomputing Theory and Methodology.
\newblock \href
  {http://dx.doi.org/https://doi.org/10.1016/j.neucom.2006.10.026}
  {\path{doi:https://doi.org/10.1016/j.neucom.2006.10.026}}.

\bibitem{Slowik:08}
A.~Slowik, M.~Bialko, Training of artificial neural networks using differential
  evolution algorithm, in: 2008 Conference on Human System Interactions, 2008,
  pp. 60--65.
\newblock \href {http://dx.doi.org/10.1109/HSI.2008.4581409}
  {\path{doi:10.1109/HSI.2008.4581409}}.

\bibitem{Arka:17}
A.~Ghosh, S.~Das, S.~S. Mullick, R.~Mallipeddi, A.~K. Das, A switched parameter
  differential evolution with optional blending crossover for scalable
  numerical optimization, Applied Soft Computing 57 (2017) 329--352.
\newblock \href {http://dx.doi.org/https://doi.org/10.1016/j.asoc.2017.03.003}
  {\path{doi:https://doi.org/10.1016/j.asoc.2017.03.003}}.

\bibitem{Kundu:13}
R.~Kundu, R.~Mukherjee, S.~Das, A.~V. Vasilakos, Adaptive differential
  evolution with difference mean based perturbation for dynamic economic
  dispatch problem, in: 2013 IEEE Symposium on Differential Evolution (SDE),
  2013, pp. 38--45.
\newblock \href {http://dx.doi.org/10.1109/SDE.2013.6601440}
  {\path{doi:10.1109/SDE.2013.6601440}}.

\bibitem{Jin:05}
Y.~Jin, J.~Branke, Evolutionary optimization in uncertain environments-a
  survey, IEEE Transactions on Evolutionary Computation 9~(3) (2005) 303--317.
\newblock \href {http://dx.doi.org/10.1109/TEVC.2005.846356}
  {\path{doi:10.1109/TEVC.2005.846356}}.

\bibitem{Fitzpatrick:88}
J.~M. Fitzpatrick, J.~J. Grefenstette, Genetic algorithms in noisy
  environments, Machine learning 3~(2) (1988) 101--120.

\bibitem{Miller:96}
B.~L. Miller, D.~E. Goldberg, Genetic algorithms, selection schemes, and the
  varying effects of noise, Evol. Comput. 4~(2) (1996) 113^^e2^^80^^93131.
\newblock \href {http://dx.doi.org/10.1162/evco.1996.4.2.113}
  {\path{doi:10.1162/evco.1996.4.2.113}}.

\bibitem{Sano:00}
Y.~Sano, H.~Kita, I.~Kamihira, M.~Yamaguchi, Online optimization of an engine
  controller by means of a genetic algorithm using history of search, in: 2000
  26th Annual Conference of the IEEE Industrial Electronics Society. IECON
  2000. 2000 IEEE International Conference on Industrial Electronics, Control
  and Instrumentation. 21st Century Technologies, Vol.~4, 2000, pp. 2929--2934
  vol.4.
\newblock \href {http://dx.doi.org/10.1109/IECON.2000.972463}
  {\path{doi:10.1109/IECON.2000.972463}}.

\bibitem{Iacca:12}
G.~Iacca, F.~Neri, E.~Mininno, Noise analysis compact differential evolution,
  International Journal of Systems Science - IJSySc 43 (2012) 1248--1267.
\newblock \href {http://dx.doi.org/10.1080/00207721.2011.598964}
  {\path{doi:10.1080/00207721.2011.598964}}.

\bibitem{Mininno:10}
E.~Mininno, F.~Neri, A memetic differential evolution approach in noisy
  optimization, Memetic Computing 2 (2010) 111--135.
\newblock \href {http://dx.doi.org/10.1007/s12293-009-0029-4}
  {\path{doi:10.1007/s12293-009-0029-4}}.

\bibitem{Nabi:10}
M.~N. Omidvar, X.~Li, Z.~Yang, X.~Yao, Cooperative co-evolution for large scale
  optimization through more frequent random grouping, in: IEEE Congress on
  Evolutionary Computation, 2010, pp. 1--8.
\newblock \href {http://dx.doi.org/10.1109/CEC.2010.5586127}
  {\path{doi:10.1109/CEC.2010.5586127}}.

\bibitem{Li:13}
X.~Li, K.~Tang, M.~N. Omidvar, Z.~Yang, K.~Qin, H.~China, Benchmark functions
  for the {CEC} 2013 special session and competition on large-scale global
  optimization, gene 7~(33) (2013) 8.

\bibitem{Wolpert:97}
D.~Wolpert, W.~Macready, No free lunch theorems for optimization, IEEE
  Transactions on Evolutionary Computation 1~(1) (1997) 67--82.
\newblock \href {http://dx.doi.org/10.1109/4235.585893}
  {\path{doi:10.1109/4235.585893}}.

\bibitem{Gayathri:01}
G.~Gopalakrishnan, B.~S. Minsker, D.~E. Goldberg, Optimal sampling in a noisy
  genetic algorithm for risk-based remediation design, Journal of
  Hydroinformatics 5 (2001) 11--25.
\newblock \href {http://dx.doi.org/10.1061/40569(2001)94}
  {\path{doi:10.1061/40569(2001)94}}.

\end{thebibliography}

\end{document}